\documentclass[journal,twoside,web]{ieeecolor}
\usepackage{generic}
\usepackage{cite}
\usepackage{amsmath,amssymb,amsfonts}
\usepackage{algorithmic}
\usepackage{graphicx}
\usepackage{algorithm,algorithmic}
\usepackage{hyperref}
\hypersetup{hidelinks=true}
\usepackage{textcomp}
\usepackage{tabularx}
\usepackage{colortbl}

\begin{document}
% \title{Towards Clinical XAI in Multimodal Vision-Language Models: ROI-Guided Perturbation Framework for Explainable Medical Image Segmentation}
\title{XAI-CLIP: ROI-Guided Perturbation Framework for Explainable Medical Image Segmentation in Multimodal Vision-Language Models}

\author{Thuraya Alzubaidi$^1$, Sana Ammar$^1$, Maryam Alsharqi$^2$, Islem Rekik$^3$, Muzammil Behzad$^{1,4,*}$
\\
$^1$King Fahd University of Petroleum and Minerals, Saudi Arabia
\\
$^2$Massachusetts Institute of Technology, US
\\
$^3$Imperial College London, UK
\\
$^4$KFUPM-SDAIA Joint Research Centre for Artificial Intelligence, Saudi Arabia
\thanks{$^*$ Corresponding author. Email: muzammil.behzad@kfupm.edu.sa}
}

\maketitle

\begin{abstract}
Medical image segmentation is a critical component of clinical workflows, enabling accurate diagnosis, treatment planning, and disease monitoring. However, despite the superior performance of transformer-based models over convolutional architectures, their limited interpretability remains a major obstacle to clinical trust and deployment. Existing explainable artificial intelligence (XAI) techniques, including gradient-based saliency methods and perturbation-based approaches, are often computationally expensive, require numerous forward passes, and frequently produce noisy or anatomically irrelevant explanations. To address these limitations, we propose XAI-CLIP, an ROI-guided perturbation framework that leverages multimodal vision-language model embeddings to localize clinically meaningful anatomical regions and guide the explanation process. By integrating language-informed region localization with medical image segmentation and applying targeted, region-aware perturbations, the proposed method generates clearer, boundary-aware saliency maps while substantially reducing computational overhead. Experiments conducted on the FLARE22 and CHAOS datasets demonstrate that XAI-CLIP achieves up to a 60\% reduction in runtime, a 44.6\% improvement in dice score, and a 96.7\% increase in Intersection-over-Union for occlusion-based explanations compared to conventional perturbation methods. Qualitative results further confirm cleaner and more anatomically consistent attribution maps with fewer artifacts, highlighting that the incorporation of multimodal vision-language representations into perturbation-based XAI frameworks significantly enhances both interpretability and efficiency, thereby enabling transparent and clinically deployable medical image segmentation systems.
\end{abstract}


\section{Introduction}
\label{sec:introduction}
\IEEEPARstart{M}{edical} image segmentation is a fundamental part of clinical workflows, allowing clinicians to assess anatomical boundaries and pathological structures for decision making in critical areas such as oncology, neurology, and cardiology \cite{deFauw2018}. Recent segmentation models that leverage transformer-based architectures far surpass traditional CNNs by effectively modeling long-range dependencies and capturing both global and local features, making them particularly effective for medical image segmentation tasks \cite{ma2024segment}. However, the transparency of their decision-making processes remains ambiguous. A recent study found that about 76\% of medical imaging competitions failed to provide labeling instructions, underscoring the lack of clarity in how AI models are trained and validated, particularly for complex modalities like multiparametric MRI and CT scans \cite{raedsch2023labelling}. This makes interpretability challenges a significant barrier for clinical adoption. Consequently, there is a need for explainable AI frameworks that can enhance transparency and unlock the black-box nature of these models \cite{Lapuschkin2019}, allowing clinicians to understand and trust AI decisions and deploy such models responsibly.

In this context, the current medical XAI methods face two key challenges. Gradient-based approaches, such as, Grad-CAM, generate saliency maps efficiently but occasionally highlight low-level features misaligned with anatomical structures \cite{DBLP:journals/corr/abs-1710-11063}. On the other hand, perturbation-based methods iteratively modify inputs to quantify feature importance \cite{Abe2025}, yet their exhaustive image-wide sampling incurs high computational costs as they apply perturbations even to regions outside of anatomically relevant structures, requiring hundreds of forward passes for a single explanation.

\begin{figure}[htbp]
    \centering
    \includegraphics[width=\columnwidth]{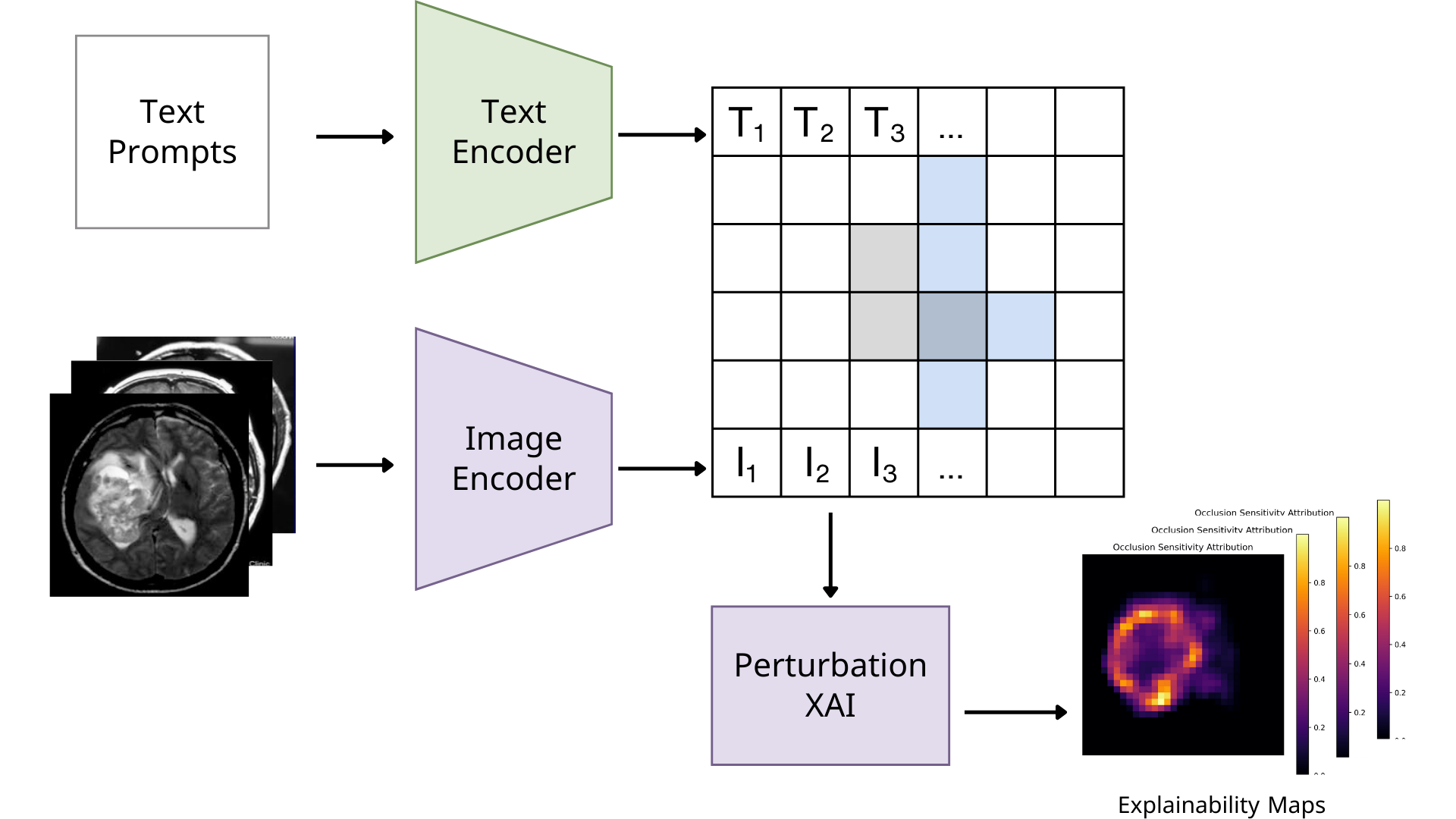}
    \caption{Overview of the XAI-CLIP framework. Text and image inputs are encoded using vision and text encoders. A joint embedding space guides region-restricted perturbation-based XAI, producing anatomically aligned explainability maps.}
    \label{fig:XAI-CLIP-overview}
\end{figure}

To address these gaps, we propose XAI-CLIP, an ROI-guided perturbation framework leveraging contrastive vision-language  modeling as shown in Fig \ref{fig:XAI-CLIP-overview}. By constraining perturbations to anatomically salient regions using the multimodal embeddings, our framework minimizes computational overhead while ensuring clinically meaningful explanations. Specifically, our framework utilizes a multimodal architecture that embeds visual and textual information in a shared representational space. The model enables zero-shot classification by measuring visual-semantic alignment between images and textual descriptions. These features can be used to extract clinically significant relationships, producing more meaningful saliency maps that align with anatomical structures and pathological regions of interest.

% The primary segmentation model used throughout the experimentation was MedSAM (Medical Segment Anything Model) \cite{ma2024segment}, a transformer-based architecture built upon the foundational Segment Anything Model (SAM) \cite{kirillov2023segment}, designed for universal medical image segmentation in different modalities such as MRI, CT, and ultrasound. Our evaluation of XAI-CLIP encompasses both qualitative and quantitative metrics, testing on samples from both FLARE22 \cite{ma2023unleashing} and CHAOS \cite{kavur2021chaos} datasets inferred on MedSAM.

The primary segmentation backbone employed in this study is MedSAM (Medical Segment Anything Model) \cite{ma2024segment}, a transformer-based framework derived from the Segment Anything Model (SAM) \cite{kirillov2023segment} and adapted for general-purpose medical image segmentation across multiple imaging modalities, including MRI, CT, and ultrasound. In this work, MedSAM serves as a representative high-performing segmentation model to evaluate the proposed explainability framework rather than as a central contribution of the study. The focus of the investigation remains on assessing the effectiveness and generality of the proposed XAI-CLIP approach in enhancing the interpretability of medical image segmentation outputs. To this end, both qualitative and quantitative analyses are conducted using samples drawn from established benchmark datasets, namely FLARE22 \cite{ma2023unleashing} and CHAOS \cite{kavur2021chaos}, which provide diverse anatomical structures and imaging characteristics suitable for evaluating explanation quality.

% The explanatory results of XAI-CLIP were compared against perturbation-based methods, and a comprehensive runtime analysis was conducted across two computational environments, Apple M1 Max MPS and NVIDIA RTX 4070 GPU, revealing that XAI-CLIP achieves up to 60\% speedup compared to conventional perturbation approaches through its region-guided methodology. Visual assessment of generated saliency maps demonstrates that XAI-CLIP produces significantly cleaner, less noisy explanations with improved boundary definition and reduced artifacts compared to existing methods.

The explanatory performance of XAI-CLIP is systematically compared against existing perturbation-based explainability methods to examine its ability to generate anatomically meaningful and visually coherent attribution maps. Rather than emphasizing experimental configurations, the discussion centers on the methodological advantages introduced by region-guided perturbations, which enable more focused and semantically aligned explanations. The proposed framework demonstrates a consistent ability to reduce unnecessary perturbations and suppress redundant  activations, leading to clearer saliency maps with improved boundary localization and fewer visual artifacts. We present detailed quantitative evaluations and computational efficiency analyses, conducted under different hardware configurations,  later in the paper, where the proposed method is shown to offer substantial runtime improvements over conventional perturbation strategies without compromising explanation fidelity.

\section{Related Work}

% \subsection{Medical Image Segmentation Models}
% With the advent of SAM, a foundation model for image segmentation that enables zero-shot generalization through a promptable architecture \cite{he2023accuracy, huang2023segment}, consisting of a powerful image encoder, a flexible prompt encoder, and a lightweight mask decoder, numerous research efforts have been dedicated to adapting it for medical imaging purposes \cite{mazurowski2023segment}.

With the emergence of the Segment Anything Model (SAM), a large-scale foundation model for image segmentation that enables prompt-driven and zero-shot generalization \cite{he2023accuracy, huang2023segment}, significant research attention has been directed toward its adaptation for medical imaging applications \cite{mazurowski2023segment}. SAM’s architecture, composed of a high-capacity image encoder, a flexible prompt encoder, and a lightweight mask decoder, provides a generalizable framework that can be repurposed across diverse segmentation tasks. Its promptable design has motivated extensive exploration into leveraging foundation models for medical image segmentation, where data scarcity and modality heterogeneity remain persistent challenges.

% MedSAM \cite{ma2024segment, mazurowski2023segment} is one of the prominent adaptations of SAM through fine-tuning of 1.57 million image-mask pairs spanning 10 imaging modalities and more than 30 cancer types. This enables the model to cover modalities from CT to dermoscopy without task-specific weights. On nasopharyngeal cancer MRI, MedSAM attains a median DSC of 87.8\% (IQR 85.0--91.4\%), outperforming SAM by 52.3\%, U-Net by 15.5\%, and DeepLabV3+ by 22.7\%. However, the bounding box prompting of MedSAM struggles with branching or overlapping vascular structures where arteries and veins share a single box.

As an extension of SAM, MedSAM \cite{ma2024segment, mazurowski2023segment} represents a notable adaptation that is tailored to the medical domain through large-scale fine-tuning on approximately 1.57 million image-mask pairs spanning over ten imaging modalities and more than thirty cancer types. This broad training strategy enables MedSAM to generalize across modalities ranging from computed tomography to dermoscopy without relying on task-specific model variants. Empirical evaluations demonstrate strong segmentation performance across multiple benchmarks, with substantial improvements over conventional convolutional and transformer-based baselines. Nevertheless, the reliance on bounding-box prompts introduces limitations in anatomically complex scenarios, particularly in cases involving branching or overlapping vascular structures, where a single bounding box may inadequately capture fine-grained spatial distinctions.

% Most recently, SemiSAM+ was proposed as a framework designed to improve semi-supervised medical image segmentation by linking SAM with task-specific segmentation networks \cite{zhang2025semisam}. The system leverages SAM-generated priors to guide pseudo-label propagation and applies refinement through downstream architectures without retraining the frozen SAM encoder. On the left atrium dataset, it reports a Dice score of 90.14\% using only five labeled samples, outperforming prior approaches like Mean Teacher and CPS in similar low-annotation settings \cite{zhang2025semisam}. This approach illustrates an ongoing move in medical image segmentation toward integrating foundation models with minimal retraining, in line with other recent efforts that rely on lightweight adapters for clinical adaptation \cite{wu2023medsamadapter, ruan2024vmunet}. Nonetheless, SemiSAM+ remains sensitive to the quality of its prompts, and its segmentation accuracy can decline when anatomical boundaries are less clearly defined. Additionally, recent studies have explored knowledge distillation techniques to further enhance semi-supervised segmentation performance \cite{huang2025knowsam}.

More recently, SemiSAM+ has been introduced as a semi-supervised medical image segmentation framework that integrates SAM with task-specific segmentation networks \cite{zhang2025semisam}. Rather than retraining the frozen SAM encoder, the method exploits SAM-generated priors to guide pseudo-label propagation, followed by refinement using downstream architectures. This strategy demonstrates strong performance in low-annotation regimes, achieving competitive Dice scores with only a small number of labeled samples on benchmark datasets \cite{zhang2025semisam}. SemiSAM+ reflects a broader trend toward coupling foundation models with lightweight adaptation mechanisms, aligning with recent approaches that employ adapter-based tuning for clinical deployment \cite{wu2023medsamadapter, ruan2024vmunet}. Despite its advantages, the framework remains sensitive to prompt quality, with segmentation accuracy degrading in regions with ambiguous or weak anatomical boundaries. Complementary efforts have also explored knowledge distillation strategies to further enhance semi-supervised segmentation performance \cite{huang2025knowsam}.

% Ruan et al.\ (2024) proposed VM-UNet, the first pure state space-model network for medical image segmentation, replacing convolutional/self-attention blocks with Vision State Space modules from Mamba \cite{ruan2024vmunet}. Their architecture uses 27.4M parameters and 4.11G FLOPs, 20\% lighter than counterparts while maintaining accuracy. VM-UNet achieves 89.03\% Dice on ISIC17, 89.71\% on ISIC18, and 81.08\% Dice with 19.21mm HD95 on Synapse, outperforming Swin-UNet.

Beyond foundation model adaptations, alternative architectural paradigms have been investigated to improve segmentation efficiency and scalability. Ruan et al.\ proposed VM-UNet, a segmentation network based entirely on state space models, replacing conventional convolutional and self-attention blocks with Vision State Space modules derived from Mamba \cite{ruan2024vmunet}. The resulting architecture significantly reduces computational complexity while maintaining competitive accuracy, achieving strong Dice scores across multiple datasets and outperforming transformer-based counterparts such as Swin-UNet. This work highlights the potential of state space models as an efficient alternative to attention-based architectures in medical image segmentation.

% While VM-UNet focused on SSM architecture, Wu et al.\ took a different approach with Med-SAM, an adapter that grafts bottlenecks onto frozen SAM for medical image segmentation \cite{wu2025medsa}. Their Space-Depth Transpose enables 3D adaptation of the 2D ViT encoder; with 75\% overlap prompts, it reaches 89.8\% Dice and 2.18mm HD95 on BTCV, surpassing Swin-UNetr with only 13M tunable weights. However, Med-SAM shows prompt sensitivity, with Dice decreasing from 89.8\% to 87.6\% when prompt overlap drops to 50\%.

% Additionally, UltraLight VM-UNet introduces a parallel Vision Mamba approach, achieving competitive performance with only 0.049M parameters and 0.060 GFLOPs on skin lesion datasets, highlighting the potential for lightweight SSM-based models in medical image segmentation \cite{wu2024ultralight}.

In a related direction, Wu et al.\ introduced Med-SAM, an adapter-based framework that augments a frozen SAM backbone with lightweight bottleneck modules to enable effective medical image segmentation \cite{wu2025medsa}. The proposed Space-Depth Transpose mechanism facilitates the extension of a 2D vision transformer encoder to volumetric data, enabling efficient 3D adaptation with a limited number of trainable parameters. While Med-SAM achieves strong segmentation accuracy with substantially fewer tunable weights compared to transformer-based baselines, its performance remains sensitive to prompt design, with notable degradation observed under reduced prompt overlap. Complementing these efforts, UltraLight VM-UNet further demonstrates the feasibility of highly compact state space-based architectures, achieving competitive performance on skin lesion segmentation tasks with minimal parameter and computational budgets \cite{wu2024ultralight}. Together, these studies underscore ongoing efforts to balance segmentation accuracy, computational efficiency, and adaptability in modern medical image segmentation models.

\subsection{Explainable AI in Medical Image Analysis}

% Explainable AI has gained traction in medical imaging due to demands for transparency in clinical decision-making. Fontes et al.\ \cite{fontes2024application} provided a domain-focused taxonomy mapping XAI methods to imaging modalities and clinical tasks. Their review extracted nine studies from 2021--2023, synthesizing prototype, retrieval, counterfactual, and GAN approaches. The review lacks quantitative aggregation and restricts inclusion to English papers from a narrow timeframe.
Explainable artificial intelligence has received increasing attention in medical imaging, driven by the need for transparency and accountability in clinical decision-making. In this regard, Fontes et al.\ \cite{fontes2024application} presented a domain-specific taxonomy that systematically categorizes XAI techniques according to imaging modalities and associated clinical tasks. Their review analyzed nine representative studies published between 2021 and 2023, encompassing a range of explainability paradigms, including prototype-based, retrieval-based, counterfactual, and generative adversarial approaches. While the study provides valuable conceptual organization of existing methods, it does not include quantitative performance aggregation and limits its scope to English-language publications within a relatively narrow temporal window.

% Addressing this limitation, Hou et al.\ \cite{hou2024self} surveyed self-explainable AI (S-XAI) techniques for medical image analysis, examining over 200 studies from 2018 to 2024. The authors organized works along input, model, and output explainability axes. The survey documented a six-fold growth in publications, with 50\% of XAI work targeting medical imaging. The paper compiled a resource table of over 70 datasets but reported performance statistics without harmonized conditions.
Addressing limitations in earlier surveys, Hou et al.\ \cite{hou2024self} presented a comprehensive review of self-explainable artificial intelligence (S-XAI) methods for medical image analysis, encompassing more than 200 studies published between 2018 and 2024. The authors structured the literature along three complementary dimensions, input-level, model-level, and output-level explainability, and reported a substantial increase in research activity, with medical imaging accounting for approximately half of all XAI-related publications. In addition, the survey compiled an extensive resource table covering over 70 publicly available datasets. However, while performance metrics were reported across studies, the absence of standardized experimental settings limits direct quantitative comparison.

% Building on these taxonomies, Iglesias et al.\ \cite{iglesias2025mocvae} introduced MOC-VAE with a SPEA2 optimizer to locate minimal latent perturbations forcing misclassification. Their approach achieved 80.56\% overlap between saliency maps and pathology regions, outperforming alternatives by 7.23--56.39 points. However, the model's diagnostic performance (0.61 accuracy) lags behind state-of-the-art chest X-ray models.
Building upon these conceptual frameworks, Iglesias et al.\ \cite{iglesias2025mocvae} proposed MOC-VAE, an explanation method that employs a SPEA2 optimizer to identify minimal perturbations in the latent space that induce misclassification. The proposed approach achieved a high degree of overlap between generated saliency maps and annotated pathological regions, surpassing several competing methods by a notable margin. Nevertheless, the underlying model exhibits relatively modest diagnostic accuracy compared to state-of-the-art chest X-ray classifiers, which constrains its clinical applicability despite improved explainability. From a theoretical standpoint, Abe and Asai \cite{abe2025towards} introduced a perturbation-based explanation framework that quantifies local input sensitivity via Jacobian matrix analysis. Their method enables instance-specific, data-free explanations that align with key clinical interpretability requirements. While the framework is mathematically well-founded, the study does not provide quantitative validation on medical imaging benchmarks, leaving its practical effectiveness largely unexplored.

% From a theoretical perspective, Abe and Asai \cite{abe2025towards} proposed a perturbation-based explanation framework quantifying local input sensitivity through Jacobian matrix computation. Their work enables data-free, instance-specific explanations meeting clinical requirements. The study presents no quantitative benchmarks on medical imaging datasets.

% Shifting to collaborative paradigms, Mastoi et al.\ \cite{mastoi2025interpretable} designed a federated learning workflow training GoogLeNet on 7,023 brain MRI images while preserving data privacy. Their ten-client approach yielded 94.24\% accuracy across four tumor classes. They embedded Grad-CAM in the inference pipeline but performed no quantitative XAI assessment.
In a collaborative learning context, Mastoi et al.\ \cite{mastoi2025interpretable} developed an interpretable federated learning pipeline for brain tumor classification, training a GoogLeNet architecture on distributed MRI data while preserving patient privacy. The proposed ten-client framework achieved high classification accuracy across multiple tumor categories. Although Grad-CAM visualizations were incorporated into the inference process, the work did not include a quantitative evaluation of explanation quality or consistency, limiting insight into the reliability of the generated saliency maps.

% In specialty applications, Yoshida et al.\ \cite{yoshida2025explainable} fine-tuned RETFound-OCT to classify OCT B-scans, using RELPROP for saliency maps. This uncovered two novel biomarkers—the ``ice pick sign'' (prevalence 32.4\%, specificity 88\%) and ``salt and pepper sign'' (prevalence 16.0\%, specificity 91\%). However, RELPROP saliency is not consistently pathologic, indicating potential over-attribution.
In domain-specific applications, Yoshida et al.\ \cite{yoshida2025explainable} fine-tuned a foundation model for optical coherence tomography (OCT) image classification and employed relevance propagation to generate saliency maps. Their analysis led to the identification of previously unreported imaging biomarkers with high specificity. However, the attribution maps produced by the explanation method were not consistently aligned with pathological regions, suggesting susceptibility to over-attribution and reduced robustness in certain cases.

% Despite these diverse approaches, current XAI methods often lack semantic alignment or region priors, operating in pixel space without leveraging semantic context from multimodal approaches.
Despite these advances across surveys, theoretical frameworks, collaborative learning, and specialty applications, many existing XAI approaches in medical imaging continue to operate primarily at the pixel level, without incorporating explicit semantic priors or anatomically meaningful region constraints. This lack of semantic grounding limits the interpretability and clinical relevance of generated explanations, particularly in complex medical imaging scenarios where contextual understanding is essential.

\subsection{Vision-Language Models in Medical Images}

% Vision-Language Models (VLMs) show significant potential in medical imaging analysis through their multimodal capabilities. In foundational research, Poudel et al.~\cite{poudel2024vlsm} evaluated VLSMs on 2D medical image segmentation across 11 diverse datasets. Their findings revealed CRIS achieved impressive 92.40\% Dice scores on BKAI while demonstrating superior generalization compared to CNN baselines, 90.23\% versus UNet's mere 36.60\% on Kvasir-SEG. This advantage extended to out-of-distribution scenarios, where CRIS maintained 64.62\% performance on ETIS when trained on ClinicDB, far surpassing UNet's 29.66\%. Despite these achievements, CLIPSeg demonstrated limited text semantics utilization, while BiomedCLIP variants unexpectedly underperformed compared to standard CLIP-based models, even with extensive pretraining on 15M medical image-text pairs.
Vision-language models (VLMs) have demonstrated significant potential for medical image analysis through joint visual-textual representation learning \cite{qiu2025refining}, exemplified by contrastive language-image pretraining (CLIP) \cite{radford2021learning}. In early foundational work, Poudel et al.\ \cite{poudel2024vlsm} systematically evaluated vision-language segmentation models on 2D medical image segmentation tasks across eleven heterogeneous datasets. Their study showed that the model achieved strong segmentation performance, reaching a dice score of 92.40\% on the BKAI dataset, while exhibiting substantially improved generalization compared to convolutional baselines, particularly in cross-dataset evaluations. Notably, the model maintained robust performance in out-of-distribution settings, significantly outperforming UNet when transferred between datasets. Despite these encouraging results, the authors observed limited utilization of textual semantics in CLIPSeg-based models, and unexpectedly weaker performance from BiomedCLIP variants, even when pretrained on large-scale medical image-text corpora \cite{poudel2024vlsm}.

% Addressing the computational efficiency challenges identified in these initial approaches, Dhakal et al.~\cite{dhakal2024vlsm} proposed VLSM-Adapter, a lightweight architecture enabling parameter-efficient fine-tuning. By requiring only 3M parameters, their method achieved comparable scores to full fine-tuning across 8 medical datasets. Their CLIPSeg Dense Adapter particularly excelled, surpassing SAN with 2.6$\times$ fewer parameters while remaining competitive with CRIS despite using $\sim$46$\times$ fewer trainable parameters. These efficiency gains, however, came with trade-offs, as performance varied significantly based on adapter placement strategy, with certain datasets still showing noticeable gaps compared to fully fine-tuned models.
To address the computational and parameter inefficiency of earlier VLM-based segmentation approaches, Dhakal et al.\ \cite{dhakal2024vlsm} introduced VLSM-Adapter, a parameter-efficient fine-tuning strategy based on lightweight adapters. Their approach reduced the number of trainable parameters to approximately three million while achieving performance comparable to full fine-tuning across multiple medical datasets. In particular, the CLIPSeg Dense Adapter variant demonstrated strong efficiency, outperforming SAN with significantly fewer parameters and remaining competitive with CRIS despite an order-of-magnitude reduction in trainable weights. However, these gains were accompanied by sensitivity to adapter placement, and performance gaps persisted on certain datasets when compared to fully fine-tuned models.

% Building on these single-model approaches, Dietlmeier et al.~\cite{dietlmeier2025vlsmensemble} developed VLSM-Ensemble techniques that strategically combined CLIP-based VLSMs with lightweight UNet architectures. Their framework introduced three ensemble variants that integrated VLSMs with low-complexity UNet through feature-space concatenation and Efficient Channel Attention layers. This combinatorial approach yielded substantial improvements, with BiomedCLIPSeg-A boosting Dice scores by 6.3\% on BKAI (0.75120~$\rightarrow$~0.81423), 4.7\% on ClinicDB, and 2.5\% on Kvasir-SEG. Yet the ensembles exhibited modality-specific inconsistencies, notably underperforming the baseline CLIPSeg-B by 2.1\% on CheXlocalize datasets.
Moving beyond single-model adaptations, Dietlmeier et al.\ \cite{dietlmeier2025vlsmensemble} proposed an ensemble-based framework that combines CLIP-based VLMs with lightweight UNet architectures. Their approach integrates feature representations through concatenation and Efficient Channel Attention mechanisms, enabling complementary strengths between multimodal and convolutional components. This ensemble strategy yielded notable improvements in segmentation accuracy across several datasets, particularly for BiomedCLIPSeg-based variants. Nevertheless, the method exhibited modality-dependent variability, with certain datasets showing reduced performance relative to baseline VLM models, highlighting challenges in achieving consistent gains across diverse imaging domains.

% While previous approaches focused on architectural modifications, Chen et al.~\cite{chen2025causalclipseg} addressed underlying model reasoning through CausalCLIPSeg, which incorporated causal intervention specifically for referring medical image segmentation, achieving 85.21 Dice and 76.90 mIoU on QaTa-COV19, outperforming LViT by 1.55\% Dice and 1.79\% mIoU, respectively. Their cross-modal decoding scheme effectively enabled text-to-pixel alignment using CLIP's text embeddings and vision encoder features, though its reliance on CLIP models trained exclusively on natural images limited understanding of specialized medical terminology.
Rather than focusing solely on architectural refinements, Chen et al.\ \cite{chen2025causalclipseg} addressed reasoning deficiencies in VLM-based segmentation by introducing CausalCLIPSeg, which incorporates causal intervention mechanisms for referring medical image segmentation. Their framework improved both dice and mIoU scores over transformer-based baselines by enforcing causal consistency between textual queries and pixel-level predictions. While the cross-modal decoding strategy enhanced text-to-pixel alignment, its reliance on vision-language models pretrained exclusively on natural images constrained its ability to fully capture specialized medical terminology and domain-specific semantics.

% For enhanced cross-modal integration, Sultan et al.~\cite{sultan2025bipvl} proposed BiPVL-Seg with bidirectional fusion and global-local contrastive alignment. This facilitated deeper information exchange between text and vision encoders while simultaneously aligning embeddings at both class and concept levels. The resulting performance gains were substantial across four medical datasets, improving DSC by +1.88\% on AMOS22 (88.21 vs 86.33) and reducing HD95 by 1.8mm. Complementing this work, Chen et al.~\cite{chen2025bivlgm} reformulated vision-language matching as a graph matching problem through Bi-VLGM, which preserved intra-modal relations and significantly improved scores on datasets like BCSS (80.96\% mDice, 70.30\% mIoU, 93.59\% mAcc)—representing gains of 2.3\% mDice and 2.6\% mIoU over ConvNeXt and DHUNet baselines.
To strengthen cross-modal interaction, Sultan et al.\ \cite{sultan2025bipvl} proposed BiPVL-Seg, a bidirectional fusion framework augmented with global-local contrastive alignment. This design enabled deeper information exchange between vision and language encoders while aligning representations at both concept and class levels. The resulting model achieved consistent performance improvements across multiple datasets, including higher dice scores and reduced Hausdorff distance. Complementarily, Chen et al.\ \cite{chen2025bivlgm} reframed vision-language correspondence as a graph matching problem through Bi-VLGM, preserving intra-modal structural relationships and yielding substantial gains over strong convolutional and hybrid baselines on histopathology benchmarks.

% Recognizing the importance of domain expertise integration, Nath et al.~\cite{nath2025vila} developed VILA-M3, incorporating expert knowledge through specialized instruction fine-tuning. Their VILA-M3-40B model outperformed Med-Gemini by an average of 9\%, achieving impressive 90.4\% accuracy on RadVQA and 92.7\% classification F1 on PathCXR. However, this domain specialization came at a cost, showing up to 23\% performance degradation on non-medical VLM benchmarks and creating dependencies on external expert models that may not always be accessible.
Recognizing the importance of domain knowledge, Nath et al.\ \cite{nath2025vila} introduced VILA-M3, which integrates expert medical knowledge via instruction-based fine-tuning. Their large-scale model achieved strong performance across medical question answering and classification benchmarks, outperforming existing medical VLMs by a significant margin. However, this specialization introduced trade-offs, including degraded performance on general-domain tasks and reliance on external expert systems, which may limit scalability and accessibility in real-world clinical settings.

% Taking a more holistic view of the field, Hartsock and Rasool~\cite{hartsock2024survey} conducted a comprehensive survey of medical VLMs specifically focused on report generation and visual question answering. Their analysis taxonomized 15 VLMs and 17 multimodal datasets while highlighting the growing importance of parameter-efficient fine-tuning strategies like LoRA in applications such as Visual Med-Alpaca and RaDialog. Their evaluation showed MedFlamingo achieved human-scored performance of 5.61 on VQA-RAD and 4.33 on Visual USMLE, though they identified concerning evaluation biases favoring radiology over other crucial domains, including pathology, ophthalmology, and surgical imaging.
From a broader perspective, Hartsock and Rasool \cite{hartsock2024survey} provided a comprehensive survey of medical VLMs with a focus on report generation and visual question answering. Their analysis categorized a wide range of models and datasets while emphasizing the growing role of parameter-efficient fine-tuning techniques such as LoRA. Although several models achieved strong performance on radiology-centric benchmarks, the authors identified notable evaluation biases and highlighted underrepresentation of other clinically important domains, including pathology and surgical imaging.

% In parallel developments relevant to segmentation tasks, Liang et al.~\cite{liang2025foundation} evaluated Vision Foundation Models for medical applications, particularly examining SAM variants that achieved up to 99\% parameter reduction through techniques like LoRASAM. Their analysis demonstrated lightweight adaptations, such as MobileSAM and TinySAM, could achieve remarkable 48.9$\times$ inference speedup without sacrificing accuracy, a critical consideration for clinical deployment. Despite these engineering achievements, they noted architectural modifications often lacked formal performance guarantees, while knowledge distillation approaches struggled with semantic misalignment between natural and medical image domains.
In parallel work related to segmentation, Liang et al.\ \cite{liang2025foundation} examined vision foundation models for medical applications, with particular attention to efficient adaptations of SAM-based architectures. Their study demonstrated that lightweight variants, enabled through techniques such as low-rank adaptation, could deliver substantial inference speedups without compromising accuracy. Despite these engineering advances, the authors noted that many architectural modifications lack formal performance guarantees, while knowledge distillation methods often suffer from semantic mismatches between natural and medical image domains.

% Despite these diverse advances across multiple technical dimensions, current medical VLMs face persistent challenges that limit clinical adoption. These include pronounced X-ray modality bias, insufficient training on diverse medical datasets, limited incorporation of spatial prompt semantics, and prohibitively high computational requirements for many healthcare settings. Perhaps most critically for clinical applications, most approaches lack robust mechanisms for saliency or semantic attribution, highlighting the urgent need for interpretability-aware frameworks that maintain computational efficiency while ensuring clinical relevance and trustworthiness in high-stakes medical decision support.
Despite significant progress across architectural design, efficiency optimization, causal reasoning, and domain adaptation, medical VLMs continue to face persistent challenges that hinder clinical adoption. These include modality bias, limited coverage of diverse medical datasets, insufficient integration of spatial and semantic prompts, and high computational demands. Most critically, existing approaches often lack robust mechanisms for interpretable saliency or semantically grounded attribution, underscoring the need for explainability-aware frameworks that balance interpretability, efficiency, and clinical relevance in high-stakes medical decision-making environments.

\subsection{CLIP in Medical Domain}

% CLIP approaches have gained traction in medical imaging due to their potential for zero-shot transfer and multimodal understanding. Zhao et al.\ \cite{zhao2025clip} provided a survey examining CLIP adaptations for medical imaging, analyzing 224 papers and establishing a three-tier taxonomy. Their benchmark showed specialized medical CLIP variants, like BiomedCLIP, achieve 100\% accuracy on seven imaging domains and 98.9\% on nine domains, outperforming PubMedCLIP by 5.8 points. The survey exhibits bias toward 2D modalities and relies on a single benchmark metric.
CLIP-based approaches have attracted growing interest in medical imaging due to their capacity for zero-shot transfer and multimodal representation learning. Zhao et al.\ \cite{zhao2025clip} presented a comprehensive survey of CLIP adaptations for medical imaging, analyzing 224 studies and proposing a three-level taxonomy to categorize methodological trends. Their benchmark analysis demonstrated that domain-specialized CLIP variants, such as BiomedCLIP, achieved near-saturated accuracy across multiple imaging domains, substantially outperforming general-domain medical CLIP models. Despite its breadth, the survey exhibits a strong emphasis on two-dimensional imaging modalities and relies predominantly on a single evaluation metric, limiting insights into robustness and cross-modal generalization.

% Exploring practical applications, Aleem et al.\ \cite{aleem2024salip} introduced SaLIP, a framework combining SAM with CLIP for zero-shot medical image segmentation. Their approach generates SAM masks, ranks regions via CLIP using GPT-generated prompts, and re-prompts SAM with selected bounding boxes. SaLIP achieved Dice scores of 0.94 on brain MRI, 0.83 on chest X-ray lung, and 0.81 on fetal head ultrasound. However, the approach fails if SAM cannot generate masks covering target anatomy, and CLIP shows limited spatial reasoning.
In applied segmentation settings, Aleem et al.\ \cite{aleem2024salip} proposed SaLIP, a zero-shot medical image segmentation framework that integrates SAM with CLIP-based region ranking. The method generates candidate masks using SAM, ranks them via vision–language similarity scores computed from automatically generated prompts, and subsequently refines segmentation through re-prompting. SaLIP demonstrated promising performance across several imaging modalities; however, its effectiveness depends critically on the ability of SAM to generate candidate masks that sufficiently cover the target anatomy. Furthermore, the limited spatial reasoning capacity of CLIP constrains its ability to reliably distinguish anatomically adjacent structures.

% To address these limitations, Khattak et al.\ \cite{khattak2024unimed} developed UniMed-CLIP, a dual encoder trained on 5.3 million pairs spanning six imaging modalities. Their multi-caption strategy improved zero-shot accuracy by 9.66 points across six datasets. UniMed-CLIP outperformed BiomedCLIP by 12.61 points and PMC-CLIP by 8.26 across 21 datasets while using one-third of BiomedCLIP's training volume. However, caption-derived supervision lacks spatial grounding.
To improve robustness and generalization across modalities, Khattak et al.\ \cite{khattak2024unimed} introduced UniMed-CLIP, a dual-encoder vision-language model trained on a large-scale, multimodal medical image-text corpus. By employing a multi-caption supervision strategy, the model achieved notable improvements in zero-shot performance across multiple datasets while requiring substantially less training data than comparable models. Although UniMed-CLIP demonstrates strong cross-dataset generalization, its reliance on caption-level supervision provides limited spatial grounding, which restricts its applicability for fine-grained localization and segmentation tasks.

% Taking another approach, Zhang et al.\ \cite{zhang2024mediclip} proposed MediCLIP, adapting CLIP for few-shot anomaly detection using learnable prompts, vision adapters, and synthetic anomaly generation. Their simulation approach raised Image AUROC on BrainMRI from 82.4\% to 94.8\% and Pixel AUROC on BUSI from 80.3\% to 89.9\% with just 16 normal samples. A BrainMRI-trained model achieved 91.3\% Image AUROC on BUSI without additional tuning. The study uses fixed hyperparameters across modalities and lacks comprehensive evaluation protocols.
Focusing on few-shot anomaly detection, Zhang et al.\ \cite{zhang2024mediclip} proposed MediCLIP, which adapts CLIP through learnable prompts, vision-side adapters, and synthetic anomaly generation. Their approach significantly improved both image-level and pixel-level anomaly detection performance under limited data regimes and demonstrated cross-dataset generalization without additional fine-tuning. Nonetheless, the use of fixed hyperparameters across heterogeneous modalities and the absence of standardized evaluation protocols limit the generalizability of the reported results.

% Combining vision-language models with segmentation, Koleilat et al.\ \cite{koleilat2025medclip} introduced MedCLIP-SAMv2, a framework that fine-tunes BiomedCLIP with DHN-NCE loss, derives M2IB saliency maps, and refines these with SAM. Their approach raised image-to-text retrieval to 84.70\% top-1 and text-to-image to 85.99\% on the ROCO benchmark. M2IB-prompted SAM achieved 77.61\% mean Dice across four modalities, 13.07 points higher than the baseline. However, significant prompt sensitivity remains, with Dice varying from 37.19\% to 77.76\% when prompt wording changes.
Integrating vision-language models directly with segmentation pipelines, Koleilat et al.\ \cite{koleilat2025medclip} introduced MedCLIP-SAMv2, which combines fine-tuned vision-language representations with SAM-based mask refinement. Their framework leverages saliency-driven prompting to guide segmentation and achieves substantial gains in both retrieval and segmentation performance across multiple modalities. However, the method remains highly sensitive to prompt formulation, with segmentation accuracy varying considerably under minor changes in textual input, underscoring ongoing challenges in achieving stable and reliable vision-language-guided segmentation in medical imaging. A detailed overview of these methods is provided in Table \ref{tab:litreview}.

\begin{table*}[t!]
\centering
\caption{Literature review summary of recent studies in medical imaging.}
\label{tab:litreview}
\footnotesize
\begin{tabularx}{\textwidth}{
p{2.6cm}   % Method
p{1.6cm}   % Authors
p{2.8cm}   % Model / Framework
p{2cm}   % Task
p{3.0cm}   % Dataset(s)
X          % Key findings (flexible, wraps)
}
\hline
\textbf{Reference} & \textbf{Authors} & \textbf{Model / Framework} & \textbf{Task} & \textbf{Dataset(s)} & \textbf{Key findings}\\
\hline

\rowcolor[gray]{0.95}
Accuracy of SAM in Medical Segmentation \cite{he2025computer} & He \textit{et~al.}\ (2023) &
SAM (semantic, point, box prompts) & Medical image segmentation &
12 public sets (e.g.\ BraTS, Kvasir, LiTS) & SAM underperforms in zero-shot medical segmentation; lacks generalization without domain adaptation.7\\ 

SAM for Medical Images \cite{Huang_2024} & Huang \textit{et~al.}\ (2023) &
SAM + medical‑tuned prompts & Medical image segmentation &
BraTS, ACDC, Prostate, CHAOS & Prompt tuning improves SAM, but still trails task-specific models on medical segmentation.
\\

\rowcolor[gray]{0.95}
Experimental Study on SAM for Medical Images \cite{Mazurowski_2023} & Mazurowski \textit{et~al.}\ (2023) &
SAM (point \& box prompts) & Medical image segmentation &
19 public sets (e.g.\ spine MRI, hip X‑ray) & IoU: 0.11–0.87\\

Segment Anything in Medical Images \cite{ma2024segment} & Ma \textit{et~al.}\ (2024) &
MedSAM (fine‑tuned SAM) & Medical image segmentation &
1.57 M image–mask pairs across 10 modalities & Outperforms SOTA in 146 tasks\\

\rowcolor[gray]{0.95}
SemiSAM+: Semi‑Supervised Segmentation via Foundation Models \cite{zhang2025semisam} & Zhang \textit{et~al.}\ (2025) &
SemiSAM+ (SAM + specialist) & Semi‑supervised segmentation &
Two public sets, one clinical set & Significant improvement with limited annotations\\

Medical SAM Adapter for Segmentation \cite{wu2024ultralight} & Wu \textit{et~al.}\ (2023) &
Medical SAM Adapter (Med‑SA) & Medical image segmentation &
17 tasks across various modalities & Outperforms SOTA while updating only 2\,\% of parameters\\

\rowcolor[gray]{0.95}
SAM‑Induced Prompt Distillation for Semi‑Supervised Segmentation \cite{huang2025knowsam} & Huang \textit{et~al.}\ (2025) &
KnowSAM (SAM + LPS + SKD) & Semi‑supervised segmentation &
Multiple public datasets & Outperforms SOTA in semi‑supervised settings\\

VILA‑M3: Expert‑Guided VLMs \cite{nath2025vilam3enhancingvisionlanguagemodels} & Nath \textit{et~al.}(2025)  &
VILA‑M3 & Multi‑task med.\ VLM &
RadQA, MIMIC‑CXR, SLAKE, PathVQA, CheXpert, ChestX‑ray14 & Avg.\ +9\,\% over Med‑Gemini \\

\rowcolor[gray]{0.95}
VLMs for Report Generation \& VQA (Review) \cite{hartsock2024survey} & Hartsock \textit{et~al.} (2024) &
Various VLMs & Report generation, VQA &
MIMIC-CXR, IU-Xray, MedICaT, VQA-RAD, PathVQA, VQA-Med 2020 & Performance varies across tasks; challenges in clinical validity and interpretability \\

Vision Foundation Models in Medical Imaging \cite{liang2025foundation} & Liang \textit{et~al.} (2025) &
Various (e.g.\ SAM, ViT) & Medical image tasks &
Multiple public sets & Near‑SOTA accuracy with few labels; robustness unresolved \\

\rowcolor[gray]{0.95}
CLIP in Medical Imaging - Survey \cite{Zhao_2025} & Zhao \textit{et~al.} (2025) &
CLIP & Various medical tasks &
Many public datasets & Contrastive pre‑training enables strong zero‑shot transfer \\

Test‑Time Adaptation with SaLIP \cite{aleem2024testtimeadaptationsalipcascade} & Aleem \textit{et~al.} (2024) &
SaLIP (SAM + CLIP) & Zero‑shot segmentation &
Brain MRI, Chest X‑ray, Fetal US & Dice: 63.46\,\% (brain), 50.11\,\% (lung), 30.82\,\% (fetal head) \\

\rowcolor[gray]{0.95}
UniMed‑CLIP for Diverse Modalities \cite{khattak2024unimed} & Khattak (2025) \textit{et~al.} &
UniMed‑CLIP & Multi‑modal imaging \cite{khattak2024unimed} &
UniMed (5.3 M pairs) & +12.61\,\% over BiomedCLIP \\

MediCLIP: Few‑shot Anomaly Detection \cite{zhang2024mediclip} & Zhang \textit{et~al.} (2024) &
MediCLIP (learnable prompts + adapters) & Few‑shot anomaly detection &
CheXpert, BrainMRI, BUSI & AUROC: +10\,\% over SOTA \\

\rowcolor[gray]{0.95}
MedCLIP‑SAMv2 for Text‑Driven Segmentation \cite{koleilat2025medclip} & Koleilat \textit{et~al.}  (2025) &
MedCLIP‑SAMv2 & Text‑driven segmentation &
Breast US, Brain MRI, Lung X‑ray/CT & Dice: +13.07\,\% (zero‑shot), +11.21\,\% (weakly supervised) \\
\hline
\end{tabularx}
\end{table*}

\section{The Propose XAI-CLIP Method}

\subsection{Preprocessing}
Our preprocessing pipeline transforms raw medical images into standardized formats suitable for our proposed pipeline. The process begins with standard input handling, including directory reading and format verification. Images are resized to 224×224 pixels using area interpolation to maintain aspect ratio, followed by grayscale conversion using the BT.601 luminance standard when necessary \cite{ronneberger2015unet}. One of the core contribution of our pipeline is the adaptive contrast enhancement technique that adjusts processing parameters based on the input image characteristics.

We first separate the foreground from the background using intensity thresholding:
\begin{equation}
\text{Mask}(x,y)=
\begin{cases}
1, & \text{if } I(x,y) < T_{\mathrm{bg}}\\[4pt]
0, & \text{otherwise}
\end{cases},
\label{eq:background-mask}
\end{equation}
where $I(x,y)$ is pixel intensity at position $(x,y)$, and $T_{\mathrm{bg}}$ is the background threshold.

We first identify background regions and exclude them from the enhancement process to avoid amplifying noise and non-informative pixels. A binary region-of-interest (ROI) mask is then applied to isolate foreground anatomical structures, enabling targeted and anatomically meaningful preprocessing. We subsequently perform percentile-based contrast stretching, in which the effective intensity range is adaptively determined from the foreground pixel distribution. Specifically, the lower and upper intensity bounds are computed using the 5th and 95th percentile values of the foreground region histogram, providing robust normalization while mitigating the influence of extreme outliers. By restricting enhancement operations to foreground pixels within this adaptive range, we improve structural visibility and ensure consistent contrast enhancement across diverse medical imaging modalities and acquisition conditions as:
\begin{equation}
F = \bigl\{\, I(x,y) \,\bigm|\, \text{Mask}(x,y)=0 \,\bigr\}.
\label{eq:foreground-set}
\end{equation}
The lower and upper intensity bounds are then determined using the 5th and 95th percentiles of the foreground distribution:
\begin{equation}
\begin{aligned}
p_{\text{low}}  &= \operatorname{percentile}\!\bigl(F, 5\bigr), \\
p_{\text{high}} &= \operatorname{percentile}\!\bigl(F, 95\bigr).
\end{aligned}
\label{eq:percentile-bounds}
\end{equation}
Finally, linear contrast stretching is applied to each pixel as:
\begin{equation}
I_{\text{str}}(x,y) = 
\begin{cases} 
0 & I(x,y) \le p_{\text{low}} \\
\frac{255 \cdot [I(x,y) - p_{\text{low}}]}{p_{\text{high}} - p_{\text{low}}} & p_{\text{low}} < I(x,y) < p_{\text{high}} \\
255 & I(x,y) \ge p_{\text{high}}
\end{cases},
\label{eq:linear-stretch}
\end{equation}
This approach provides robust normalization across varied imaging conditions without relying on fixed global thresholds.
For local contrast enhancement, we employ Contrast Limited Adaptive Histogram Equalization (CLAHE) \cite{mishra2021clahe} with an 8×8 tile grid. The histogram clipping mechanism prevents noise amplification in homogeneous regions:
% ---------- b) Clip histogram (drop-in fragment) ----------
\begin{equation}
\begin{aligned}
H'_T(b) &= \min\bigl(H_T(b), C\bigr) + \frac{1}{256} \sum_{i=1}^{256} \max\bigl(0, H_T(i) - C\bigr) \\
&\text{subject to } C = \frac{2.0 \cdot N_{\text{pixels}}}{256}
\end{aligned}
\label{eq:clahe-clip}
\end{equation}
The cumulative distribution function is computed as:
\begin{equation}
\text{CDF}(b)=\sum_{k=1}^{b} H'_{T}(k).
\label{eq:clahe-cdf}
\end{equation}
% ---------- End fragment ---------------------------------
This is followed by histogram equalization as:
% ---------- d) Equalize (CLAHE output) ----------

\begin{equation}
I_{\text{CLAHE}}(x,y)=
\Bigl[
  \frac{\text{CDF}\!\bigl(I(x,y)\bigr) - \text{CDF}_{\min}}
       {\text{CDF}_{\max} - \text{CDF}_{\min}}
  \times 255
\Bigr].
\label{eq:clahe-equalize}
\end{equation}
To preserve the natural appearance of background regions while enhancing diagnostic features, we selectively apply the processing:

\begin{equation}
I_{\text{enhanced}}(x,y)=
\begin{cases}
I_{\text{CLAHE}}(x,y)     & \text{if } \text{Mask}(x,y)=0,\\[4pt]
I_{\text{original}}(x,y)  & \text{otherwise}.
\end{cases}
\label{eq:enhanced-output}
\end{equation}

\begin{figure}[t!]
\centering

% -------- Row 1 --------
\begin{minipage}{0.31\columnwidth}
    \centering
    \includegraphics[width=\linewidth]{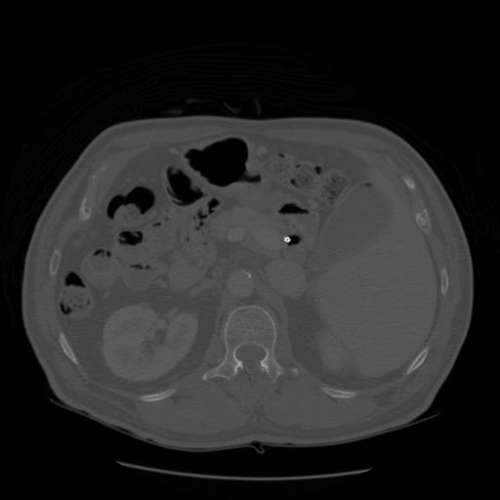}
\end{minipage}\hfill
\begin{minipage}{0.31\columnwidth}
    \centering
    \includegraphics[width=\linewidth]{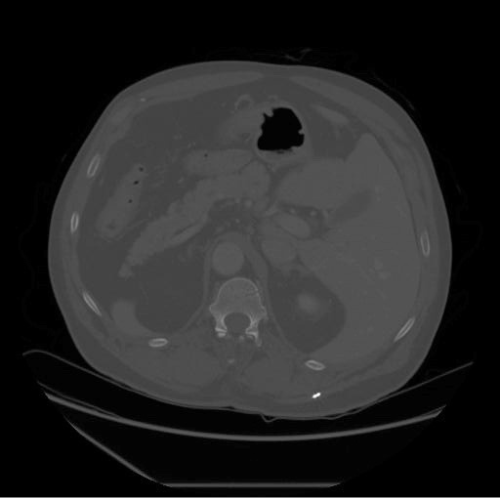}
\end{minipage}\hfill
\begin{minipage}{0.31\columnwidth}
    \centering
    \includegraphics[width=\linewidth]{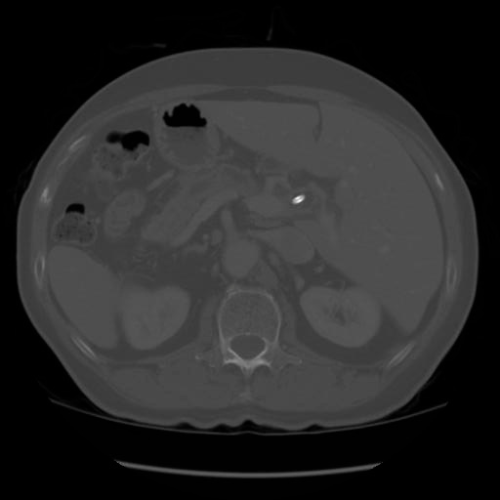}
\end{minipage}

\vspace{0.6em}
% -------- Row 2 --------
\begin{minipage}{0.31\columnwidth}
    \centering
    \includegraphics[width=\linewidth]{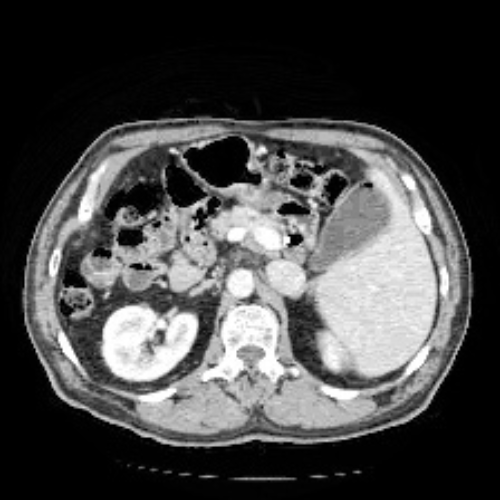}
\end{minipage}\hfill
\begin{minipage}{0.31\columnwidth}
    \centering
    \includegraphics[width=\linewidth]{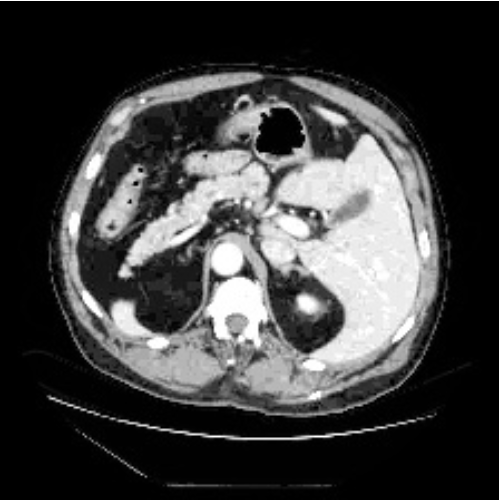}
\end{minipage}\hfill
\begin{minipage}{0.31\columnwidth}
    \centering
    \includegraphics[width=\linewidth]{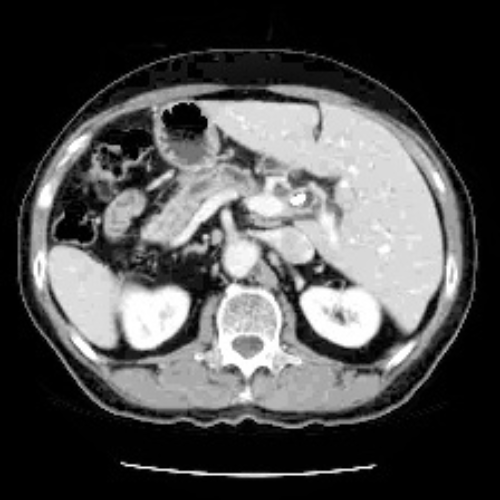}
\end{minipage}
\caption{Representative MRI slices from the FLARE22 dataset \cite{ma2023unleashing}, showing images before preprocessing (top row) and after preprocessing (bottom row) across three samples.}
\label{fig:ct_side_by_side}
\end{figure}

As shown in Fig. \ref{fig:ct_side_by_side}, we ensure that contrast enhancement is applied dynamically and adapts to each individual image by leveraging the underlying grayscale intensity distribution, rather than imposing uniform adjustments across the entire dataset. The enhanced output is subsequently validated and converted to an 8-bit representation to ensure numerical stability and compatibility with downstream processing. The pipeline supports parallel batch processing with consistent configuration settings, while allowing key parameters, such as the background intensity threshold, percentile bounds for contrast stretching, CLAHE clip limit \cite{stimper2019mclahe}, and tile grid size, to be tuned for specific datasets. Artifact suppression is achieved through smooth histogram redistribution, bilinear interpolation between tiles, and edge-aware processing strategies. This preprocessing design is particularly well suited for low-contrast medical images, especially CT and MRI scans, where preserving subtle anatomical details is essential for reliable diagnosis and accurate segmentation in subsequent deep learning pipelines.

\subsection{Perturbation-Based XAI}

% Perturbation-based XAI methods operate on a fundamental principle: systematically altering input images and measuring the resulting changes in model output. These methods analyze how modifications to specific regions of an image affect the model's predictions, thereby identifying which image features contribute most significantly to the model's decision-making process.
% In our research, we examined three perturbation-based XAI techniques: LIME (Local Interpretable Model-agnostic Explanations) \cite{aldughayfiq2023explainable}, RISE (Randomized Input Sampling for Explanation) \cite{highton2024rise3d}, and Occlusion Sensitivity analysis \cite{gandomkar2022occlusion}. While numerous other perturbation-based methods exist in the literature, these three represent diverse approaches with distinct strengths and limitations. For our experimentation, we applied these XAI methods to the MedSAM model \cite{wu2023medsamadapter}, an adaptation of the Segment Anything Model (SAM) \cite{huang2025knowsam} specifically tuned for medical image segmentation. It should be noted that the choice of MedSAM was primarily for experimental purposes, as these methods evaluated are model-agnostic and could be applied to any segmentation model.
Perturbation-based XAI methods are founded on the principle of systematically modifying input images and observing the corresponding changes in model outputs to infer feature importance. By introducing controlled perturbations to localized regions of an image and measuring their impact on prediction confidence or segmentation outcomes, these approaches identify image regions that contribute most significantly to the model’s decision-making process. In this study, we focus on three representative perturbation-based XAI techniques: Local Interpretable Model-agnostic Explanations (LIME) \cite{aldughayfiq2023explainable}, Randomized Input Sampling for Explanation (RISE) \cite{highton2024rise3d}, and Occlusion Sensitivity analysis \cite{gandomkar2022occlusion}. These methods were selected to capture a range of perturbation strategies with complementary strengths and limitations. Importantly, this choice serves as a representative testbed rather than a methodological dependency, as the evaluated XAI techniques are inherently model-agnostic and can be readily applied to alternative segmentation architectures.

\subsubsection{LIME}

\begin{figure}[b!]
  \centering
  \includegraphics[width=\columnwidth]{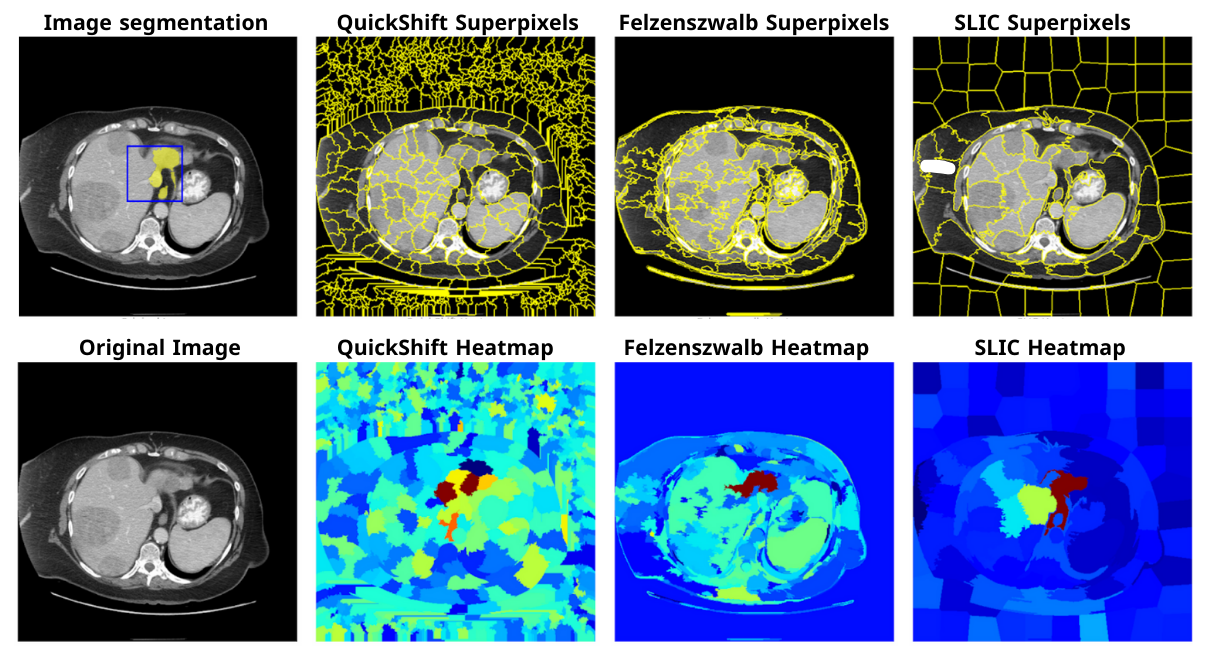}
  \caption{LIME explanations using different superpixel methods: QuickShift, Felzenszwalb, and SLIC. Top: superpixels; Bottom: Our corresponding heatmaps showing influential regions.}
  \label{fig:lime_explanations}
\end{figure}

% LIME generates explanations by approximating the complex model locally with an interpretable model \cite{aldughayfiq2023explainable}. The method divides the input image into superpixels and creates perturbed samples by randomly turning these superpixels on or off \cite{bhati2024survey}. By analyzing how these perturbations affect the model's output, LIME identifies which regions most influence the segmentation results. Our implementation tested different superpixel algorithms (QuickShift, SLIC, and Felzenszwalb), each producing distinct explanation heatmaps that highlight different aspects of the segmentation process.
LIME generates explanations by locally approximating the behavior of a complex model with an interpretable surrogate model \cite{aldughayfiq2023explainable}. The approach partitions the input image into superpixels and constructs a set of perturbed samples by randomly enabling or disabling these regions \cite{bhati2024survey}. The influence of each superpixel is then estimated by analyzing the corresponding changes in the model’s output, enabling identification of regions that most strongly affect the segmentation outcome. In our implementation, we evaluated multiple superpixel generation strategies, including QuickShift, SLIC, and Felzenszwalb, each of which yields distinct explanation heatmaps and emphasizes different spatial and structural characteristics of the segmentation process as shown in Fig \ref{fig:lime_explanations}.

\subsubsection{RISE}
\begin{figure}[t!]
\centering
  \includegraphics[width=\columnwidth]{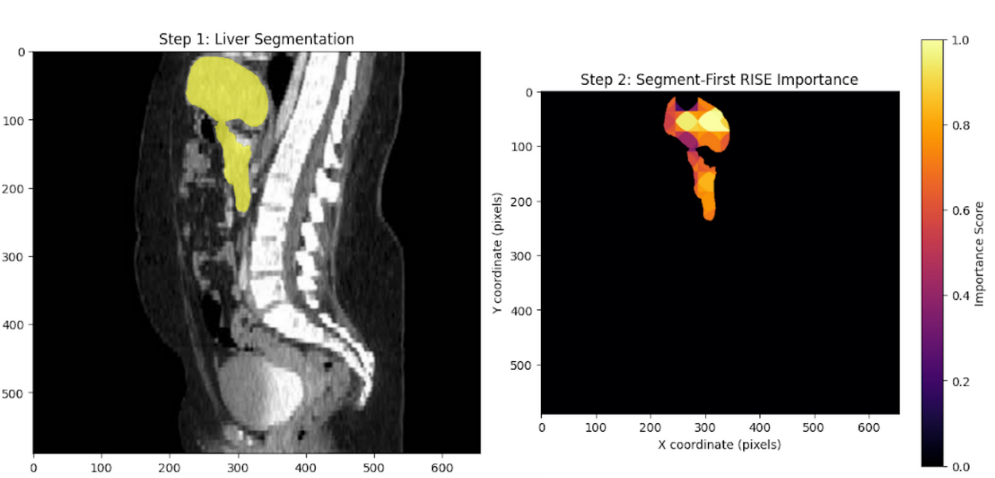}
  \caption{RISE explanation: input with segmentation (Left) and our corresponding importance map (right) showing contribution of regions to the prediction.}
  \label{fig:rise_explanation}
\end{figure}

RISE generates attribution maps through randomized masking of the input image, wherein multiple masked variants are forwarded through the model and the resulting predictions are aggregated using the corresponding masks as weights \cite{highton2024rise3d}. This stochastic sampling process estimates pixel-wise importance by correlating prediction sensitivity with mask presence. In our observations, applying RISE over the entire image space often produced diffuse and less interpretable attribution maps. Consequently, constraining the masking process to organ-specific regions yielded more localized and semantically meaningful explanations. Under this region-restricted setting, RISE consistently differentiated between internal organ structures and boundary regions, assigning higher relevance to interior anatomical areas while attenuating spurious activations along edges as shown in Fig~\ref{fig:rise_explanation}.

\subsubsection{Occlusion Sensitivity Analysis}

\begin{figure}[b!]
\centering
  \includegraphics[width=\columnwidth]{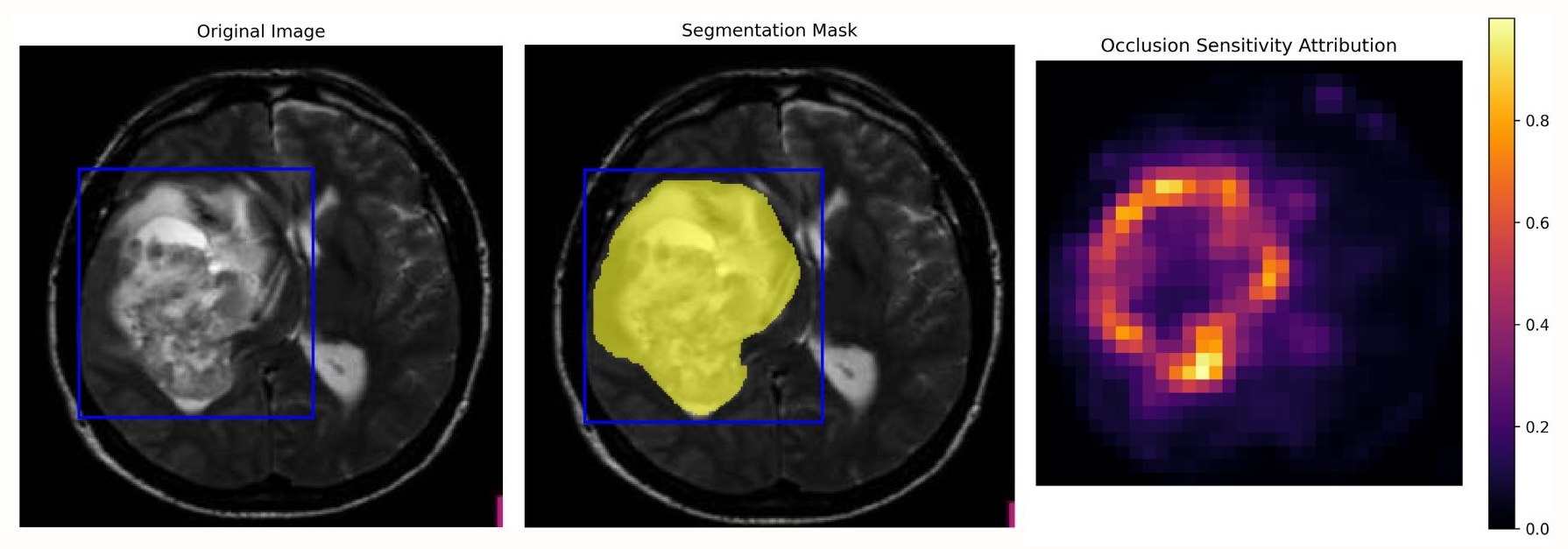}
  \caption{Occlusion explanation: original input (left), segmentation mask (middle), and our importance heatmap (right) showing regions critical to the prediction.}
  \label{fig:occlusion_explanation}
\end{figure}

% Occlusion Sensitivity systematically blocks portions of the input image with a sliding window to determine which regions most affect the segmentation outcome \cite{resta2021occlusion}. By occluding different parts of the image and measuring changes in the model's output, this method creates an importance map highlighting critical regions. This approach offers a straightforward analysis method that proved particularly suitable for multi-layer medical scans, with its simple occlusion-based technique providing intuitive explanations.
Occlusion Sensitivity assesses feature importance by systematically masking localized regions of the input image using a sliding window and measuring the resulting changes in the model’s output \cite{resta2021occlusion}. By iteratively occluding different spatial locations and quantifying their impact on the segmentation predictions, the method produces an importance map that highlights regions critical to the model’s decision process. This straightforward perturbation strategy offers intuitive and easily interpretable explanations and was found to be particularly effective for multi-layer medical imaging data, where the localized occlusion mechanism provides clear insights into the spatial dependencies learned by the segmentation model as shown in Fig \ref{fig:occlusion_explanation}.

\subsection{Anatomical Region Extraction}

% The localization of regions of interest (ROIs) was achieved using \textit{MediCLIP} (Medical Contrastive Language-Image Pretraining) \cite{zhang2024mediclip}, a CLIP-based model that aims to adapt CLIP for few-shot medical image anomaly detection \cite{zhao2025clip}, and is utilized in this study to guide explainability analysis. The model was pretrained on medical images via the \texttt{open\_clip} framework. Input images undergo preprocessing while preserving aspect ratio to maintain compatibility with CLIP's positional embedding grid. Bounding box coordinates are correspondingly rescaled and shifted to align with the new padded image space.

% The input images pass through the CLIP visual encoder to extract tokenized embeddings. \textit{Necker} processes multi-scale token outputs from CLIP to adapt them for downstream tasks, while a learned \textit{Adapter} further transforms these vision features. A \textit{PromptMaker} encodes a predefined set of prompts such as ``abnormal organ'' or ``healthy tissue.''

% For generating region importance maps, the \textit{MapMaker} module fuses visual features with prompt features to produce anomaly/importance maps. The second channel of the output specifically denotes abnormal or important regions. The importance map undergoes normalization to the range \([0, 1]\) and is smoothed using a Gaussian blur. A binary ROI mask is then generated by applying a fixed threshold (\(\geq 0.5\)). This ROI mask is stored and utilized to constrain subsequent perturbation analysis, ensuring focus on clinically relevant areas.

We perform region-of-interest (ROI) localization using MediCLIP \cite{zhang2024mediclip}, a medical adaptation of contrastive vision-language pretraining originally proposed for few-shot anomaly detection in medical images \cite{zhao2025clip}. In our framework, MediCLIP is employed as a semantic localization module to guide explainability analysis rather than as a diagnostic classifier. The model is pretrained on medical image-text pairs using the \texttt{open\_clip} framework. Input images are preprocessed with aspect ratio preservation to maintain compatibility with the positional embedding grid of the vision-language encoder, and corresponding bounding box coordinates are rescaled and shifted to align with the padded image space.

We pass the preprocessed images through the visual encoder to extract token-level embeddings, which are subsequently processed by the Necker module to aggregate multi-scale representations for downstream localization. A learned Adapter further refines these visual features, while a PromptMaker encodes a predefined set of semantic prompts, such as “abnormal organ” or “healthy tissue,” into the shared embedding space. To generate region importance maps, we fuse the adapted visual features with the prompt embeddings using the MapMaker module, producing an output in which a dedicated channel corresponds to abnormal or clinically relevant regions. The resulting importance map is normalized to the range \([0, 1]\), smoothed via Gaussian filtering, and binarized using a fixed threshold (\(\geq 0.5\)) to obtain a binary ROI mask. This mask is stored and subsequently used to constrain perturbation-based explainability analyses, ensuring that the generated explanations remain focused on clinically meaningful anatomical regions.

\subsubsection{CoOp-Based Prompt Optimization}
% Recently, utilizing CoOp (context optimization) prompt learning instead of manually designed prompts in the medical domain has gained more attention \cite{koleilat2024biomedcoop}. While manually designed prompts work well in gathering semantic information in natural images, they struggle to capture important and subtle details in medical images \cite{zhang2024mediclip}. In order to address this, CoOp prompt learning is utilized, where each prompt structure includes learnable word embeddings (tokens). In addition to introducing prompt learning instead of manual prompt designing, CoOp needs only a few labeled images for learning, which is crucial in the medical field, as annotating data can be time-consuming, and it addresses the scarcity of annotated medical images \cite{zhou2022conditional}. The prompt structure is defined as [V1][V2]...[VM][CLS], where [V] represents a learnable word embedding, M denotes the number of these learnable tokens, and [CLS] represents the class. These learnable tokens function as the continuous context that CoOp introduced, allowing the model to optimize the prompt's context during training. Prompts for each specific class are then formed by combining the learnable context vectors with the word embeddings of the class name. Then, the CLIP text encoder transforms these prompts into feature representations.

Recent work has increasingly explored the use of context optimization (CoOp) prompt learning as an alternative to manually designed prompts in medical vision-language applications \cite{koleilat2024biomedcoop}. While handcrafted prompts have proven effective for capturing semantic information in natural images, they often fail to represent the subtle and clinically relevant variations present in medical imagery \cite{zhang2024mediclip}. To address this limitation, CoOp introduces learnable prompt representations in which each prompt consists of a sequence of trainable word embeddings that are optimized during training. This formulation enables the model to adapt prompt context directly from data, rather than relying on fixed textual descriptions. Importantly, CoOp requires only a limited number of labeled samples to learn effective prompt representations, making it particularly suitable for medical imaging scenarios where annotation is costly and scarce. Formally, the prompt structure is defined as [V1][V2]...[VM][CLS], where [V] denotes learnable context tokens, M is the number of such tokens, and [CLS] represents the class label. These learnable context vectors are concatenated with the class name embeddings to form class-specific prompts, which are subsequently encoded by the vision-language model's text encoder to produce optimized textual feature representations.

\subsection{Segmentation-guided Organ Localization}
To adapt MediCLIP for multi-organ region classification, we reformulate the original anomaly detection formulation into a region-aware semantic learning framework. This transition requires architectural modifications, prompt redefinition, and optimization adjustments to enable explicit spatial reasoning over anatomically meaningful regions rather than generic anomaly localization.

\subsubsection{From Anomaly Maps to Segmentation Maps}
To repurpose the MapMaker module for multi-class organ region classification, we replace the original anomaly map generator with a U-Net architecture \cite{ronneberger2015unet} that produces dense, per-pixel predictions. Although the objective is not full organ segmentation, employing a segmentation backbone enables pixel-level classification, which is essential for accurately localizing and distinguishing spatial regions corresponding to different organs. This design allows the model to capture fine-grained spatial structure and contextual relationships between adjacent anatomical regions. In this formulation, input images are first processed by the vision encoder to extract visual embeddings, while predefined textual prompts are encoded using the text encoder. The MapMaker then projects the prompt embeddings to match the dimensionality of the visual features, concatenates the multimodal representations, and employs a U-Net decoder to generate multi-class segmentation maps. This architectural modification effectively shifts the framework from anomaly-centric detection to region-aware semantic classification, making it more suitable for organ-level localization tasks.

% To adopt the MapMaker for our task, multi-class organ region classification, the anomaly map maker was replaced with a U-Net architecture, proposed by \cite{ronneberger2015unet}, which will generate a segmentation map instead. While our task is not to segment the organs, but rather to classify the regions the organs are in, using the U-Net as a backbone helps in classifying pixels, allowing the model to localize and distinguish between regions corresponding to different organs. This spatial understanding is crucial for accurate regional classification. The image is first passed through CLIP's encoder to generate image embeddings. The predefined text prompts are also passed through CLIP's text encoder to generate text embeddings.
% Finally, the MapMaker projects prompt embeddings to match vision feature channels, concatenates them, and uses a U-Net decoder to generate a per-class segmentation map.
% This architectural change shifts the model from anomaly-based detection to region-aware classification, helping it better adapt to our task of classifying organ regions.

\subsubsection{Prompt-Aware Organ Localization}

% With the aim of using CoOp for classifying the regions each organ resides in, five main classes were introduced to address different organs and structures: Background, Liver, Spleen, Left kidney, and Right kidney. This modification allowed the model to generate predictions that align better with the goal of organ localization rather than generic anomaly detection.
To support organ-specific regional classification using context optimization, we define a set of semantic classes corresponding to Background and organs. This explicit class formulation aligns the learned representations with anatomical structures of interest and enables the model to produce spatially coherent predictions associated with distinct organ regions. By framing localization as a multi-class semantic prediction problem, the model moves beyond generic anomaly scoring and instead learns structured region-level distinctions that are consistent with anatomical priors.

\subsubsection{Loss Function Optimization}

% To optimize the loss function, we utilized a combined loss function that integrates multi-class Dice loss and cross-entropy. The Dice loss generalizes the binary Dice loss to handle multiple classes by computing a per-class Dice score and averaging across classes, which is particularly beneficial in medical image segmentation, where class imbalance is common. While Dice loss is commonly used in medical image segmentation \cite{bertels2019optimizing}, it is still relevant here because our task involves pixel-wise classification, where spatial accuracy and region-level consistency are critical. For the other component of the combined loss function, the cross-entropy loss was implemented, which promotes accurate pixel-wise classification by penalizing incorrect predictions on a per-pixel basis.
We optimize the model is performed using a composite loss function that combines multi-class dice loss with categorical cross-entropy. The dice loss extends the conventional binary formulation to the multi-class setting by computing class-wise overlap scores and averaging across all classes, making it particularly effective in medical imaging scenarios characterized by class imbalance and varying organ sizes. Although dice loss is widely used for segmentation tasks \cite{bertels2019optimizing}, it remains well suited for the proposed pixel-wise classification objective, where spatial coherence and regional consistency are critical. The cross-entropy component complements this by enforcing accurate per-pixel classification and penalizing incorrect predictions at a fine-grained level. Together, this combined objective balances region-level overlap accuracy with pixel-level discriminative learning, facilitating stable and anatomically meaningful localization. The combined dice and cross-entropy loss is given as:
\begin{equation}
\mathcal{L}_{\text{total}} = \mathcal{L}_{\text{Dice}} + \mathcal{L}_{\text{CE}},
\end{equation}
\begin{equation}
\mathcal{L}_{\text{Dice}} = 1 - \frac{2\sum_{c=1}^{C}\sum_{i=1}^{N}p_i^{(c)}g_i^{(c)}}{\sum_{c=1}^{C}\sum_{i=1}^{N}p_i^{(c)} + g_i^{(c)} + \epsilon},
\end{equation}
\begin{equation}
\mathcal{L}_{\text{CE}} = -\sum_{i=1}^{N}\sum_{c=1}^{C}g_i^{(c)}\log(p_i^{(c)}),
\end{equation}
where $C$ is the number of classes, $N$ is the number of pixels, $p_i^{(c)}$ is the predicted probability for class $c$ at pixel $i$, $g_i^{(c)}$ is the ground truth binary label for class $c$ at pixel $i$, and $\epsilon$ is a small constant to avoid division by zero.

\subsection{CLIP-guided Perturbation-based Explainability}
% The selective application of perturbation-based explainability methods addresses a key limitation: the computational intensiveness of these XAI methods. By applying these methods only to the ROI derived from CLIP, the process reduces processing time and focuses on clinically relevant areas while requiring fewer patches to analyze.
% The implementation employs MedSAM, a variant of the Segment Anything model, to produce segmentation masks. To facilitate integration with explainability frameworks, a MedSAMWrapper class encapsulates the forward logic to ensure compatibility with any custom perturbation analysis. 
% The selective occlusion sensitivity analysis operates by occluding the input image in $64 \times 64$ patches with a stride of 32. Importantly, patches are skipped unless they overlap with the binary ROI mask. For each relevant patch, segmentation is re-executed, and the Dice difference from the baseline is calculated to compute attribution. A sensitivity map is constructed by accumulating the attribution score for each patch, with the final heatmap normalized for visualization.
% This approach prioritizes efficiency, as the number of patches processed is substantially reduced compared to full-image perturbation. Additionally, the visual analysis becomes more focused and interpretable due to the guided localization within clinically relevant regions.
The selective application of perturbation-based explainability methods is designed to address a key limitation of such approaches, namely their high computational cost. By restricting perturbations to regions of interest (ROIs) identified through vision-language-guided localization, the explainability process reduces redundant computations while concentrating analysis on clinically meaningful areas. This region-guided strategy significantly decreases the number of patches that must be evaluated, leading to improved runtime efficiency without sacrificing interpretability. For implementation, we employ MedSAM, a medical adaptation of the Segment Anything Model, to generate segmentation masks that serve as the basis for attribution analysis. To ensure seamless integration with perturbation-based explainability frameworks, we introduce a MedSAMWrapper class that encapsulates the model’s forward pass and standardizes its interface for compatibility with custom perturbation routines. Selective occlusion sensitivity is then performed by occluding the input image using $64 \times 64$ patches with a stride of 32, while explicitly skipping patches that do not intersect with the binary ROI mask. For each retained patch, segmentation is recomputed and the change in Dice score relative to the unperturbed baseline is measured to quantify attribution. These attribution scores are accumulated to form a sensitivity map, which is subsequently normalized for visualization. Overall, this selective perturbation scheme substantially reduces computational overhead and yields more focused, anatomically aligned explanations by confining analysis to clinically relevant regions.

\begin{figure}[b!]
  \centering
    \includegraphics[width=\columnwidth]{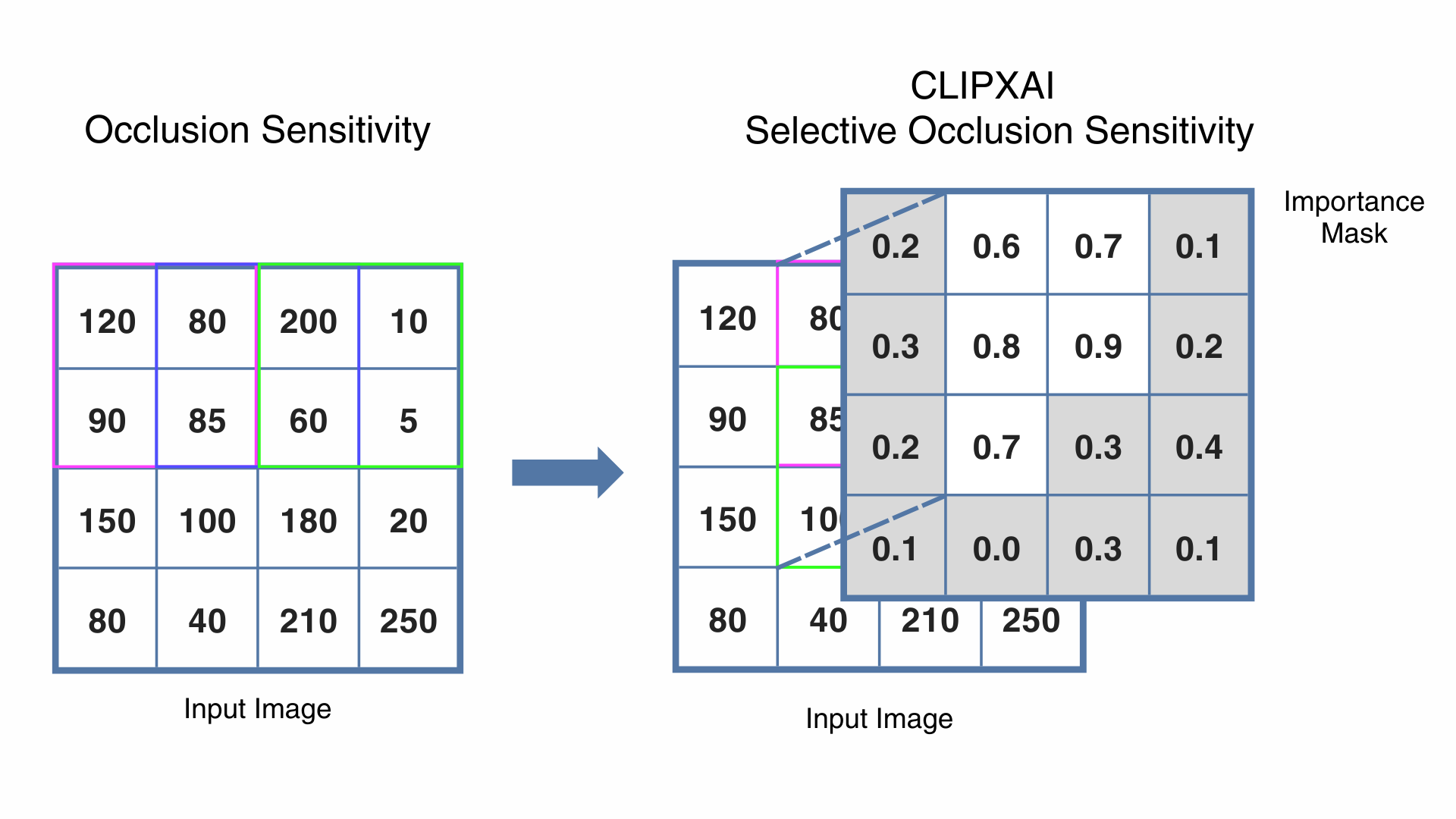}
  \caption{Comparison of standard occlusion sensitivity (left) and XAI-CLIP-based selective occlusion sensitivity (right). The proposed method incorporates an importance mask to guide occlusion toward informative regions, improving efficiency by avoiding irrelevant areas. }
  \label{fig:adapted_xai}
\end{figure}

\subsubsection{Occlusion-based Anatomical Region Extraction}

Our anatomical region extraction strategy is built upon a selective occlusion sensitivity framework that concentrates computational effort on clinically relevant anatomical structures. To achieve this, we employ a wrapper-based implementation that preserves the original segmentation model architecture while enabling seamless integration with perturbation-based explainability pipelines. This design ensures compatibility with downstream analysis modules without altering the underlying model behavior, while maintaining spatial consistency and contextual awareness during perturbation.
\begin{figure}[b]
  \centering
  \includegraphics[width=\linewidth]{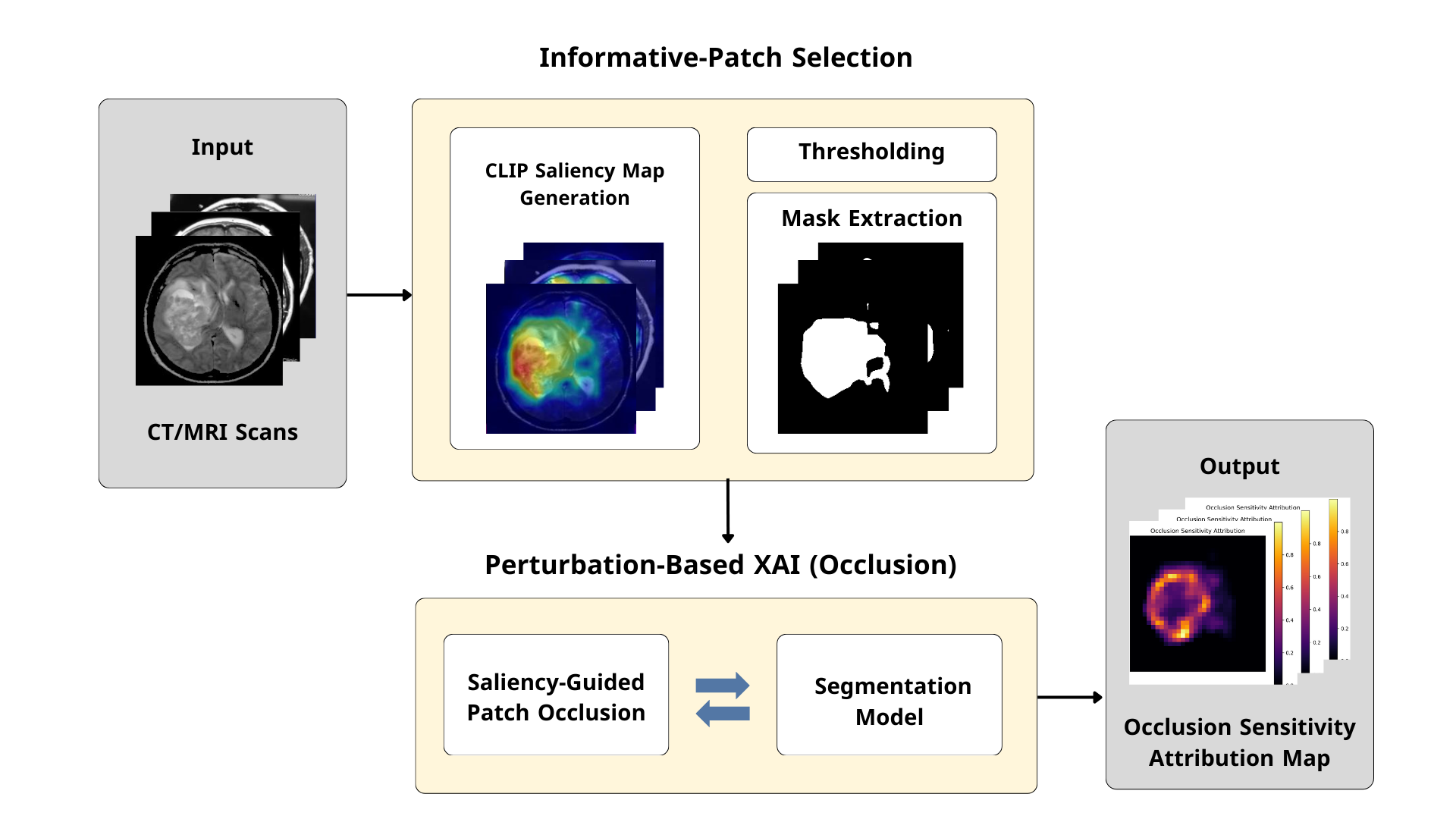}
  \caption{Overview of the occlusion sensitivity method. Saliency maps are thresholded to extract ROI masks, guiding patch-level occlusion and generating attribution maps for model interpretability. }
  \label{fig:adapted_xainew}
\end{figure}

As shown in Fig. \ref{fig:adapted_xai}, the central contribution of this framework is the introduction of ROI-guided importance masking, in which binary or grayscale region-of-interest masks defining anatomically relevant areas are first thresholded to obtain binary representations. These masks are then resized to match the model input resolution using nearest-neighbor interpolation to preserve hard anatomical boundaries and avoid interpolation-induced artifacts. The resulting ROI masks serve as spatial constraints that guide the perturbation process, ensuring that explainability analysis remains focused on meaningful anatomical regions rather than background or irrelevant areas.

Building on this representation, we implement a selective patch processing strategy based on a sliding-window occlusion scheme using $64 \times 64$ pixel patches with a stride of 32 pixels. Unlike conventional occlusion sensitivity analysis, patches are processed only if they overlap with the ROI mask, effectively filtering out non-informative regions. This selective mechanism reduces the number of evaluated patches from approximately 10,000 in full-image analysis to between 500 and 2,000 patches, depending on the size of the ROI and the applied importance threshold (e.g., retaining the top 20\% most relevant regions). As a result, the proposed strategy achieves a computational reduction of approximately 75-95\% while preserving the fidelity of the resulting explanations. The accumulated perturbation responses are aggregated into occlusion heatmaps that highlight anatomically significant regions by quantifying the impact of localized occlusions on segmentation performance, as shown in Fig. \ref{fig:adapted_xainew}.

While occlusion sensitivity analysis forms the foundation of the proposed anatomical region extraction methodology, we further investigate complementary perturbation-based explainability techniques, including LIME and RISE, to provide a broader validation of the framework across different attribution mechanisms. This multi-method evaluation enables a more comprehensive assessment of explanation consistency and robustness across diverse perturbation strategies.

\subsubsection{Analysis of Additional Perturbation Methods}

While occlusion sensitivity analysis forms the foundation of our anatomical region extraction methodology, we subsequently explored the RISE approach to provide complementary probabilistic importance estimates. The RISE implementation generates stochastic binary masks using binomial distribution sampling, producing \(N = 2000\) random masks with a configurable density parameter (\(p_{1} = 0.5\)) at a compact \(7 \times 7\) base resolution. These masks are then upsampled to the full input dimensions using bilinear interpolation. This process reduces the memory footprint by over an order of magnitude compared to full-resolution mask generation (e.g., \(\approx 1000\times\) for a \(224 \times 224\) image, whereas \(214\times\) would correspond to an \(\approx 15 \times 15\) grid) while maintaining adequate spatial precision through upsampling. where optional random shifts of the masks further mitigate grid artifacts.

When applying ROI-constrained perturbation, only pixels within the predefined regions of interest are subjected to perturbation, while pixels outside the ROI are explicitly preserved by assigning a fixed multiplier value of 1.0. This constraint ensures that attribution analysis remains focused on anatomically and clinically relevant areas while preventing spurious effects from background regions. In contrast to the deterministic sliding-window mechanism employed by occlusion sensitivity analysis, RISE adopts a stochastic perturbation strategy based on randomly generated masks, producing probabilistic attribution maps that capture expected importance through aggregation over multiple randomized perturbations.

The methodology further incorporates a dual saliency mapping strategy, in which separate importance maps are generated to capture segmentation fidelity and general ROI relevance, respectively. Independent min-max normalization is applied to each map to preserve relative importance and dynamic range across different anatomical and functional region types, enabling more nuanced interpretation of model behavior across complementary explanatory dimensions.

% Further experimenting with LIME, the implementation employs Felzenszwalb superpixel segmentation with parameters optimized for medical imaging: \texttt{scale} = 100 to balance detail preservation and computational efficiency, \texttt{sigma} = 0.5 to apply Gaussian smoothing while preserving organ boundaries, and \texttt{min\_size} = 50 to avoid fragmentation of non-clinical features. Superpixel generation is restricted to regions of interest (ROIs) by zeroing out non-ROI pixels and numbering ROI superpixels sequentially from 1 to $N$, reducing $512 \times 512$ images to approximately 150 semantic regions.
In further experiments with LIME, we employ Felzenszwalb superpixel segmentation with parameters tailored to medical imaging characteristics. Specifically, the \texttt{scale} parameter is set to 100 to balance structural detail preservation with computational efficiency, \texttt{sigma} is fixed at 0.5 to apply mild Gaussian smoothing while maintaining organ boundary integrity, and \texttt{min\_size} is set to 50 to prevent over-fragmentation of non-clinical features. Superpixel generation is constrained to predefined regions of interest by zeroing out pixels outside the ROI and assigning sequential labels to ROI superpixels. This restriction effectively reduces a $512 \times 512$ image to approximately 150 semantically meaningful regions, enabling more focused and efficient perturbation analysis.

% The perturbation strategy involves multi-region ablation, where combinations of superpixels are systematically removed, evaluating between 200 and 500 perturbation combinations per image. This enables anatomically grounded explanations by grouping pixels into coherent regions that align with vascular structures, organ sub-regions, and pathology clusters.

% A multi-scale analysis is incorporated by combining outputs from different scale parameters (50, 100, and 200), capturing features ranging from micro-calcifications to entire organ structures. Spatial coordinates are preserved throughout to facilitate integration within clinical workflows.
The perturbation strategy is based on multi-region ablation, in which combinations of superpixels are systematically removed and evaluated, with approximately 200 to 500 perturbation samples generated per image. This formulation enables anatomically grounded explanations by operating on coherent regional groupings rather than isolated pixels, allowing the resulting attributions to align with vascular patterns, organ substructures, and localized pathological regions. To further enhance robustness, a multi-scale analysis is performed by aggregating explanations derived from multiple superpixel scales (50, 100, and 200), thereby capturing features across varying spatial extents, from fine-grained micro-calcifications to larger organ-level structures. Spatial coordinates are preserved throughout the process, ensuring that the resulting attribution maps remain compatible with downstream clinical workflows and visualization systems.

\section{Experimental Setup}
\subsection{Datasets}
% For our experiments, we utilized data from three publicly available medical imaging repositories. The primary dataset employed was FLARE22 \cite{ma2023unleashing}, which served as our main source for both training and validation. For additional testing and experimental validation, we incorporated samples from the SAROS \cite{saros2023} and CHAOS \cite{kavur2021chaos} datasets.

% All three datasets consist of abdominal CT and MRI scans in NIfTI format. Rather than using the full 3D volumes, we extracted 2D axial slices for our analysis. Specifically, we selected middle axial slices, as they typically contain clear views of all relevant abdominal organs with optimal anatomical representation.

% The FLARE22 dataset provided the core of our experimental data, while the SAROS and CHAOS samples were utilized to assess the generalizability of our approach across different acquisition protocols and scanner specifications.
For experimental evaluation, we leveraged three publicly available medical imaging datasets. The FLARE22 dataset \cite{ma2023unleashing} served as the primary source for training and validation, while additional samples from the SAROS \cite{saros2023} and CHAOS \cite{kavur2021chaos} datasets were incorporated for extended testing and validation. All datasets comprise abdominal CT and MRI scans provided in NIfTI format; however, instead of processing full 3D volumes, we extracted 2D axial slices for analysis. Middle axial slices were selected, as they consistently offer clear visualization of key abdominal organs and representative anatomical context. While FLARE22 forms the core experimental dataset, the inclusion of SAROS and CHAOS enables assessment of the robustness and generalizability of the proposed approach across varying imaging protocols and scanner characteristics.

\subsection{Configurations}
% Experiments were conducted on two systems: Machine (A): An Apple M1 Max Mac Studio featuring a system-on-chip (SoC) with a 10-core CPU (8 performance cores, 2 efficiency cores), a 32-core GPU, and a 16-core Neural Engine, coupled with 32 GB of unified LPDDR5-6400 memory (512-bit bus, 400 GB/s bandwidth). The system ran macOS 14 (Sonoma) and used PyTorch 2.2 with the Metal backend, leveraging BF16 mixed precision; Machine (B): an NVIDIA RTX 4070 GPU (12 GB GDDR6X, 192-bit bus), equipped with an Intel Core i7-13700K (16 cores, 24 threads), and 32 GB of DDR5-5600 RAM, operating under Windows 11 Pro with PyTorch 2.2, CUDA 12.4, and cuDNN 9 in FP16 mixed precision.
Experiments were performed on two computational platforms to evaluate performance across heterogeneous hardware configurations. Machine (A) was an Apple M1 Max–based Mac Studio featuring a system-on-chip architecture with a 10-core CPU (8 performance and 2 efficiency cores), a 32-core GPU, and a 16-core Neural Engine, paired with 32 GB of unified LPDDR5-6400 memory (512-bit bus, 400 GB/s bandwidth). This system ran macOS 14 (Sonoma) and utilized PyTorch 2.2 with the Metal backend, operating in BF16 mixed precision. Machine (B) consisted of an NVIDIA RTX 4070 GPU with 12 GB of GDDR6X memory (192-bit bus), an Intel Core i7-13700K processor (16 cores, 24 threads), and 32 GB of DDR5-5600 RAM. This system operated under Windows 11 Pro and employed PyTorch 2.2 with CUDA 12.4 and cuDNN 9, using FP16 mixed precision.

\subsection{Evaluation Metrics}
% We quantify computational efficiency and fidelity using five measures. 
% \textbf{Wall-clock latency} \(T\) records the elapsed seconds required to process a single slice. 
% \textbf{FLOPs} denote the total floating-point operations per slice, obtained with \textit{ptflops} by summing the ROI-extraction pass and the \(N_{\text{patch}}\) MedSAM passes executed during perturbation. 
% \(N_{\text{patch}}\) is the number of occlusion patches actually evaluated, incremented within the perturbation loop. 
% The \textbf{ROI-reduction ratio} \(\rho = N_{\text{patch}} / N_{\text{patch}}^{\text{full}}\) expresses the fraction of the full patch grid that requires computation (lower \(\rho\) indicates greater pruning). 
% Finally, \textbf{Dice} and \textbf{IoU} measure the overlap between the baseline MedSAM segmentation and the output produced after selective occlusion, providing a fidelity check.
We assess computational efficiency and explanation fidelity using five complementary metrics. Wall-clock latency ($T$) measures the elapsed time required to process a single 2D slice end to end. Floating-point operations (FLOPs) quantify the computational cost per slice and are computed using \textit{ptflops} by aggregating the operations incurred during ROI extraction and the $N_{\text{patch}}$ MedSAM forward passes executed as part of the perturbation process. Here, $N_{\text{patch}}$ denotes the number of occlusion patches actually evaluated and is recorded dynamically within the perturbation loop. The ROI-reduction ratio, defined as $\rho = N_{\text{patch}} / N_{\text{patch}}^{\text{full}}$, captures the fraction of the full patch grid that requires computation, with lower values indicating more effective pruning. Finally, dice coefficient and Intersection-over-Union (IoU) are used to quantify the overlap between the baseline MedSAM segmentation and the segmentation obtained under selective occlusion, serving as fidelity measures to ensure that computational gains do not compromise explanatory reliability.

\section{Results and Discussions}
\subsection{VLM-based Results}

% Using the FLARE22 dataset, we experimented with the MediCLIP version, which we extended for multi-class organ region classification. We focused on five classes: Background, Liver, Spleen, Right kidney, and Left kidney. Utilizing the 50 labeled images provided by the FLARE22 dataset, we divided them into training and validation sets using an 80/20 split. Each sample includes corresponding segmentation masks for the target organ classes. We evaluated performance using performance-wise AUC-ROC and Dice score. As shown in Table 2, lower Dice scores were observed across most classes, which aligns with our goal of anatomical localization rather than achieving pixel-perfect segmentation accuracy.
Using the FLARE22 dataset, we conducted experiments with the MediCLIP-based model extended to support multi-class organ region classification. The classification task was defined over five anatomical categories: Background, Liver, Spleen, Right Kidney, and Left Kidney. From the 50 labeled samples provided by FLARE22, we constructed training and validation splits using an 80/20 ratio, with each sample accompanied by corresponding segmentation annotations for the target organs. The model performance was evaluated using class-wise AUC-ROC and dice score. As reported in Table~\ref{tab:dice_auc}, the observed dice scores are relatively significant across most classes, which is consistent with the objective of regional anatomical localization rather than precise, pixel-level organ segmentation.
\begin{table}[t!]
\centering
\caption{Dice Score and AUC-ROC per Class}
\label{tab:dice_auc}
\begin{tabular}{lcc}
\hline
\textbf{Class} & \textbf{Dice Score} & \textbf{AUC-ROC} \\
\hline
Background & 0.88 & 0.86 \\
Liver & 0.64& 0.97\\
Right Kidney & 0.44& 0.97\\
Left Kidney & 0.57& 0.99\\
Spleen & 0.24& 0.98 \\
Pancreas & 0.16& 0.97 \\
\hline
\end{tabular}
\end{table} 
\begin{figure}[b!]
\centering

\begin{minipage}{0.3\columnwidth}
    \centering
    \includegraphics[width=\linewidth]{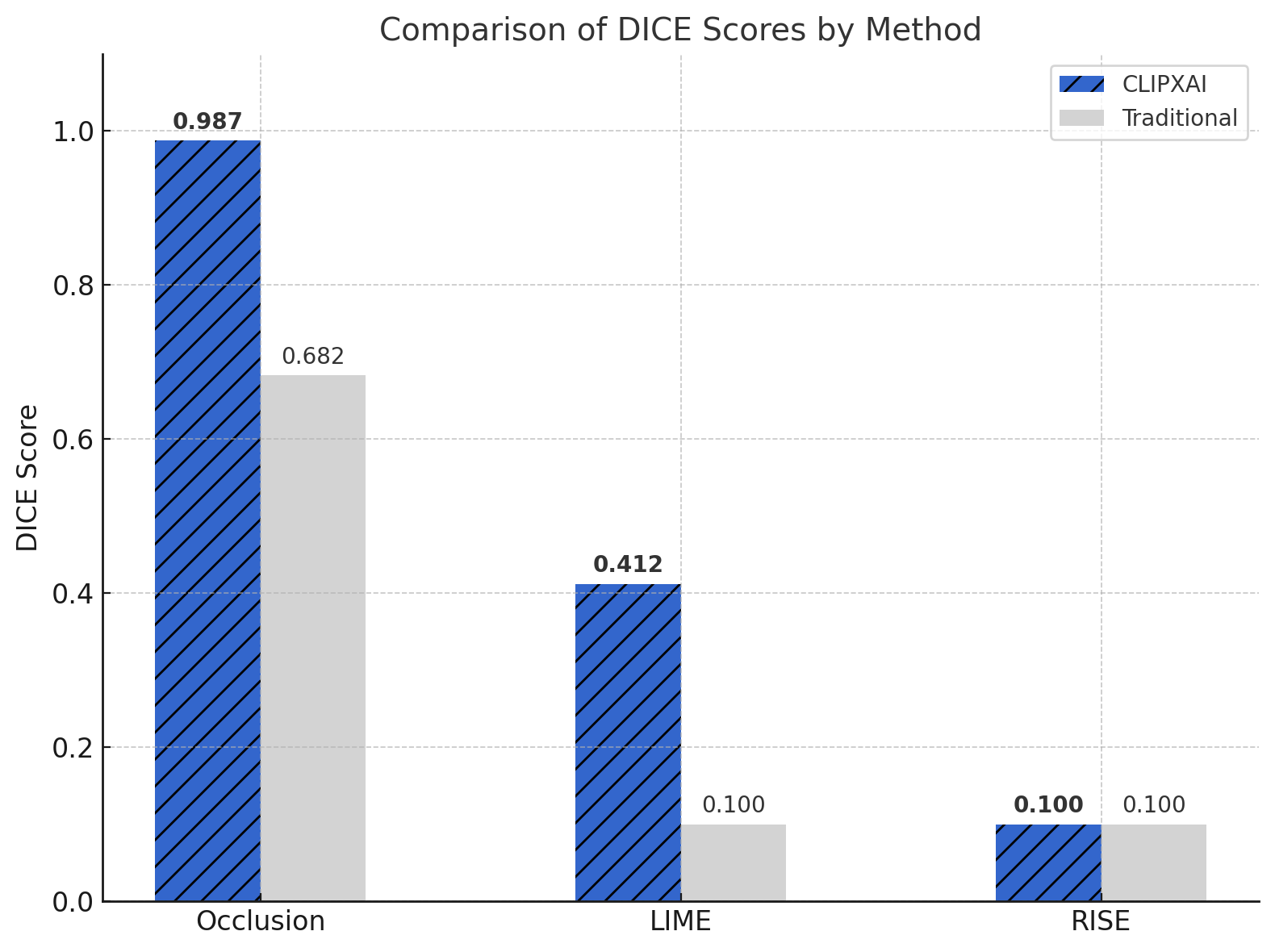}
    (a) Dice
\end{minipage}\hfill
\begin{minipage}{0.3\columnwidth}
    \centering
    \includegraphics[width=\linewidth]{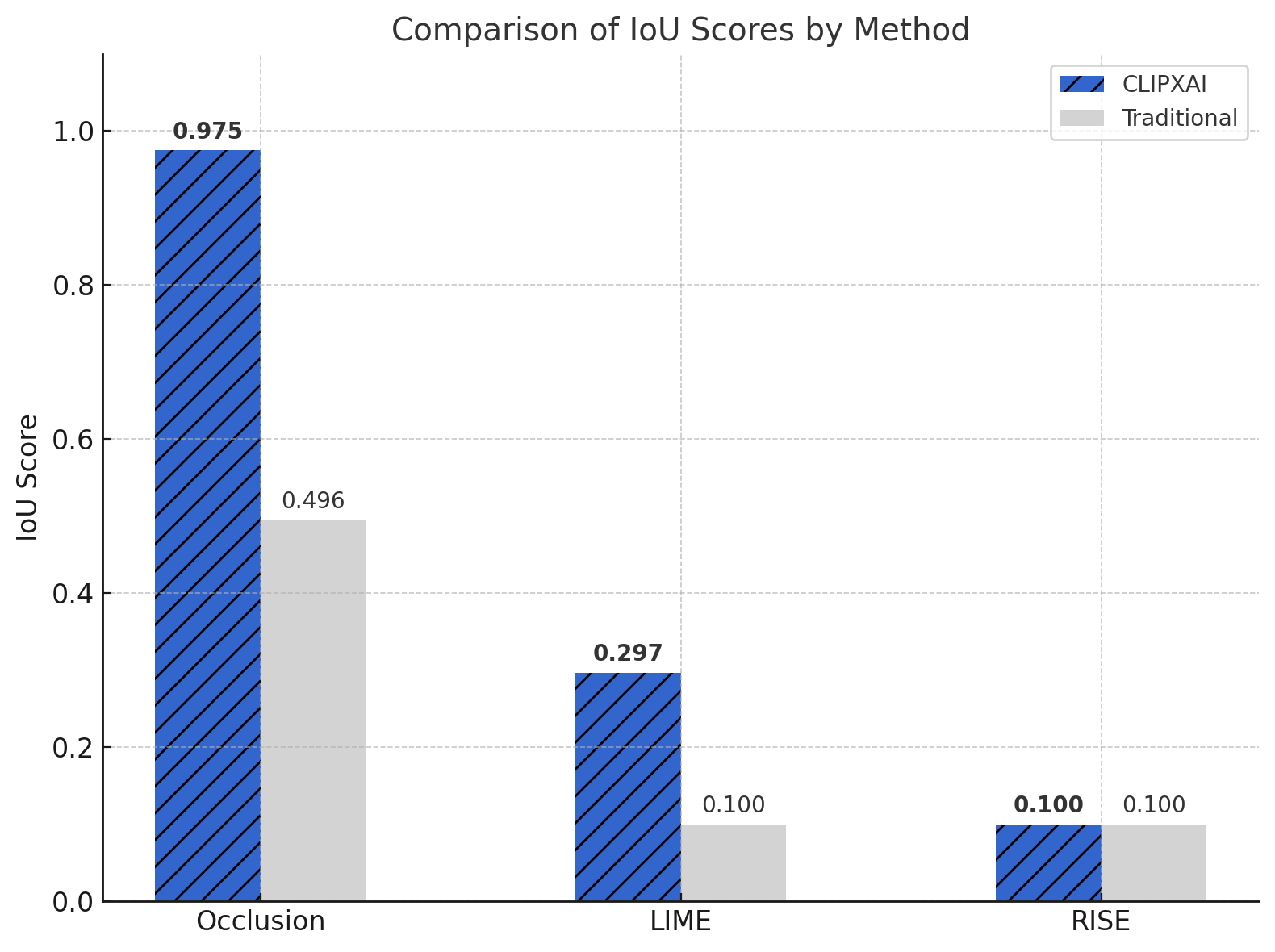}
    (b) IoU
\end{minipage}\hfill
\begin{minipage}{0.3\columnwidth}
    \centering
    \includegraphics[width=\linewidth]{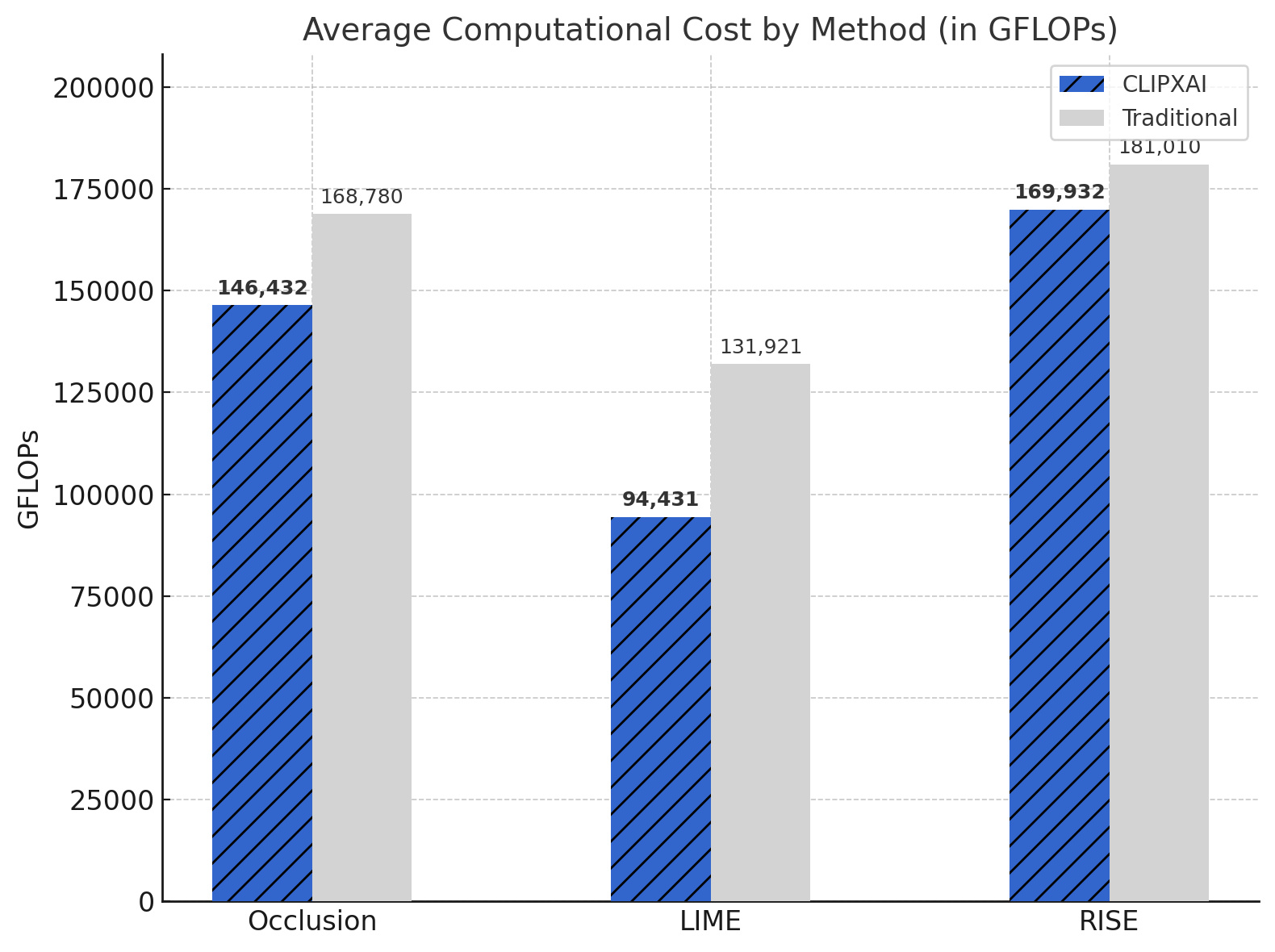}
    (c) FLOPs
\end{minipage}

\vspace{1em}

\begin{minipage}{0.6\columnwidth}
    \centering
    \includegraphics[width=\linewidth]{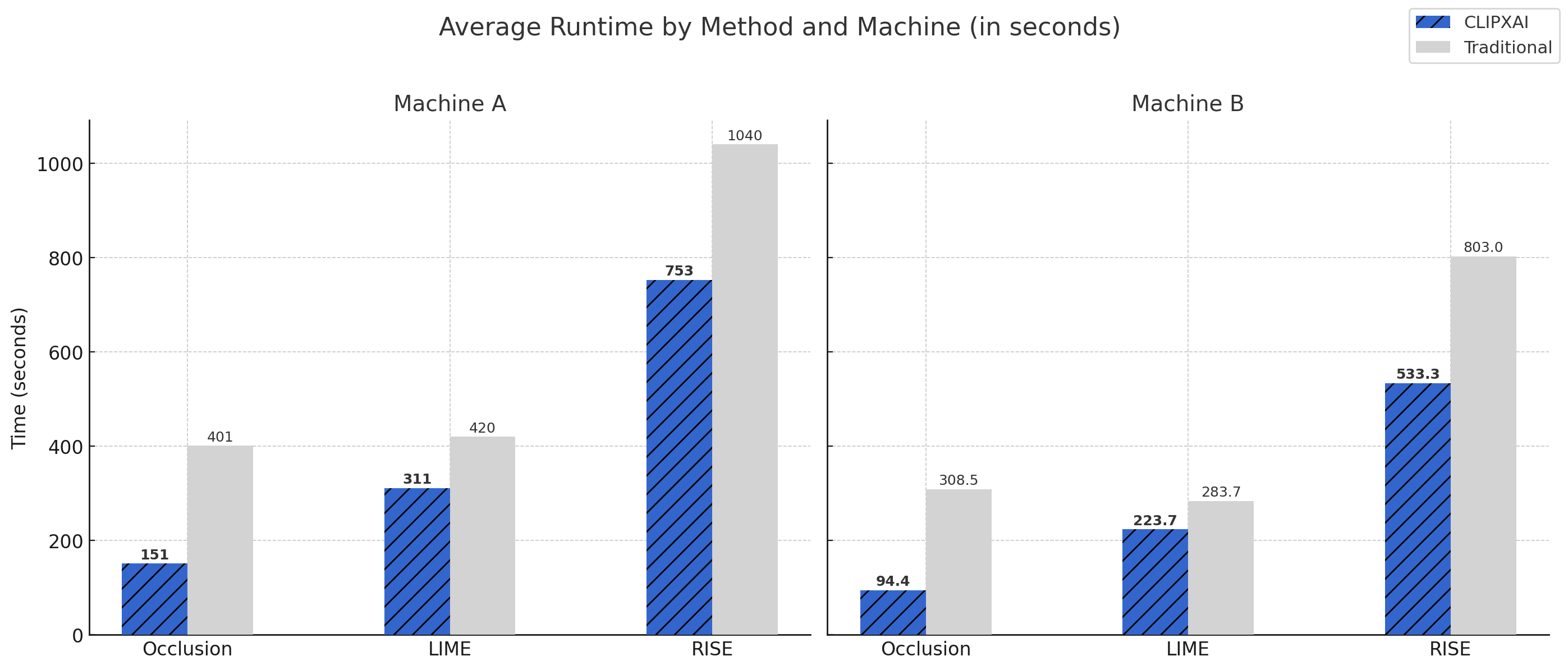}
    (d) Runtime
\end{minipage}

\caption{Quantitative comparison of explainability performance and computational efficiency: (a) Dice score, (b) IoU, (c) FLOPs, and (d) runtime across machines.}
\label{fig:barplots_compact}
\end{figure}

\begin{figure*}[t!]
\centering
\includegraphics[width=\linewidth]{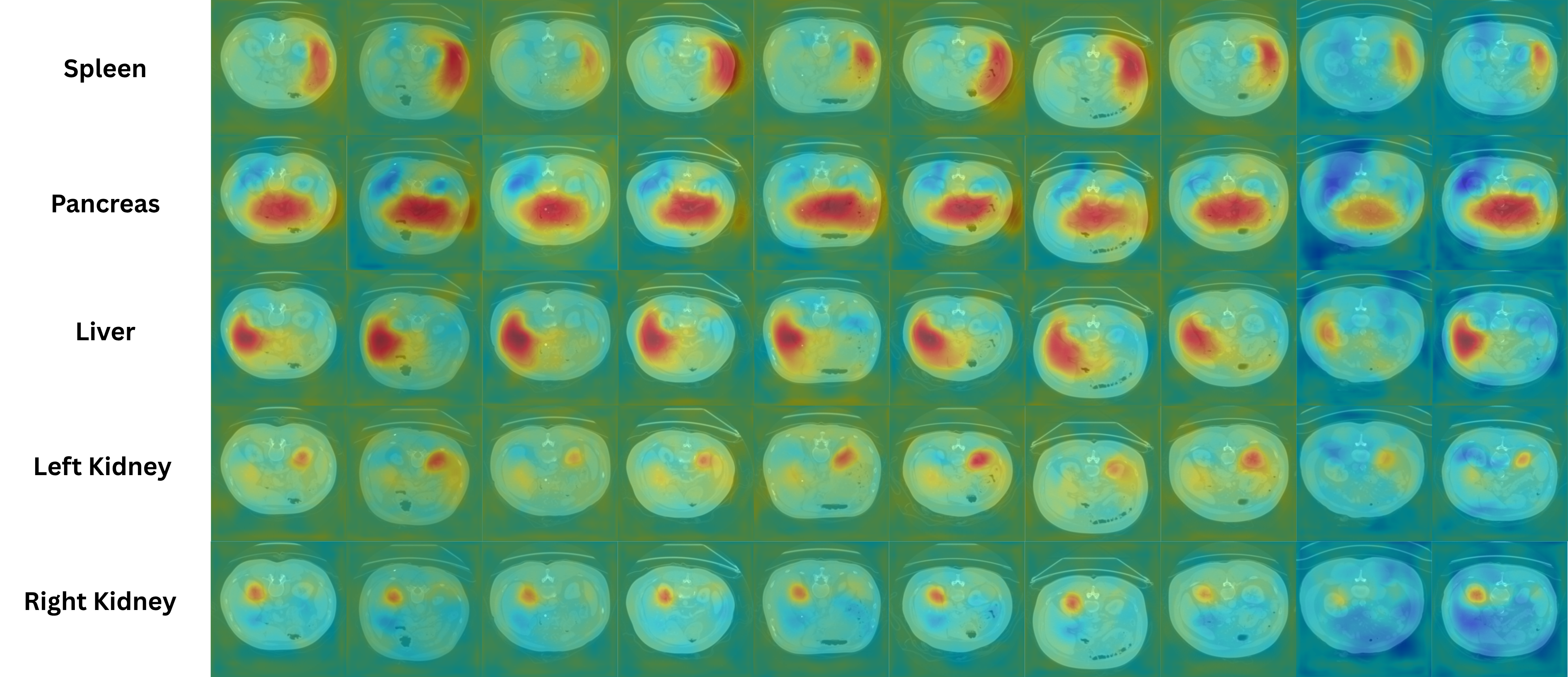}
\caption{Results on FLARE22 dataset.}
\label{fig:segmentationMap}
\end{figure*}

Similarly, as shown in Fig. \ref{fig:segmentationMap}, the model was able to locate five different organs correctly, showing that our approach of combining a U-Net-based segmentation backbone with CoOp prompt learning achieved accurate region-wise classification of organs. This architecture, therefore, is valuable for generating perturbation-based explainability.

\subsection{Perturbation-based Explainability}
% XAI-CLIP demonstrated substantial computational savings across all perturbation methods tested. Specifically, the XAI-CLIP-Occlusion approach achieved remarkable runtime reductions of 250 seconds on Hardware A and 214 seconds on Hardware B, corresponding to efficiency improvements of $62.3\%$ and $69.4\%$, respectively. Similarly, integrating XAI-CLIP using LIME as a base demonstrated considerable resource optimization, requiring only $94{,}431$ GFLOPs and representing a $28.4\%$ reduction from the baseline implementation. The RISE-based XAI-CLIP variant also achieved $27$--$34\%$ runtime savings, accompanied by an $11{,}078$ GFLOP reduction ($6.1\%$ decrease).

As shown in Fig. \ref{fig:barplots_compact}, XAI-CLIP consistently delivers substantial computational efficiency gains across all evaluated perturbation-based explainability methods. In particular, the XAI-CLIP-Occlusion variant achieves pronounced runtime reductions of 250 seconds on Machine A and 214 seconds on Machine B, corresponding to efficiency improvements of 62.3\% and 69.4\%, respectively. When integrated with LIME, XAI-CLIP further demonstrates significant computational savings, reducing the total cost to 94,431 GFLOPs, which represents a 28.4\% reduction relative to the baseline implementation. The RISE-based XAI-CLIP variant similarly exhibits notable efficiency gains, achieving 27-34\% reductions in runtime alongside a decrease of 11,078 GFLOPs (6.1\%), underscoring the effectiveness of ROI-guided perturbation in reducing computational overhead without compromising explanatory capability.

\begin{figure}[b!]
\centering
\includegraphics[width=\linewidth]{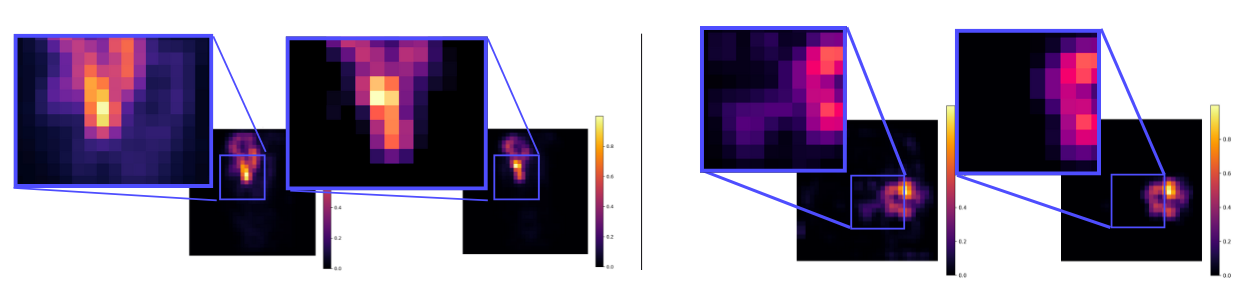}
\caption{Occlusion-sensitivity heat-map of the model output.}
\label{fig:occlusion_fullwidth}
\end{figure}

Similarly, as shown in Table \ref{tab:clip_comparison}, using occlusion sensitivity as the underlying perturbation strategy for XAI-CLIP resulted in the most pronounced performance gains among all evaluated methods. Quantitatively, this configuration achieved improvements of 44.6\% in Dice coefficient and 96.7\% in Intersection-over-Union relative to conventional occlusion-based explanations. These gains are further supported by qualitative assessment, which reveals substantially reduced noise, sharper boundary delineation, and improved spatial coherence in the generated saliency maps when compared to baseline approaches as demonstrated in Fig. \ref{fig:occlusion_fullwidth}.

\begin{figure}[b!]
\centering

\begin{minipage}{0.48\columnwidth}
    \centering
    \includegraphics[width=\linewidth]{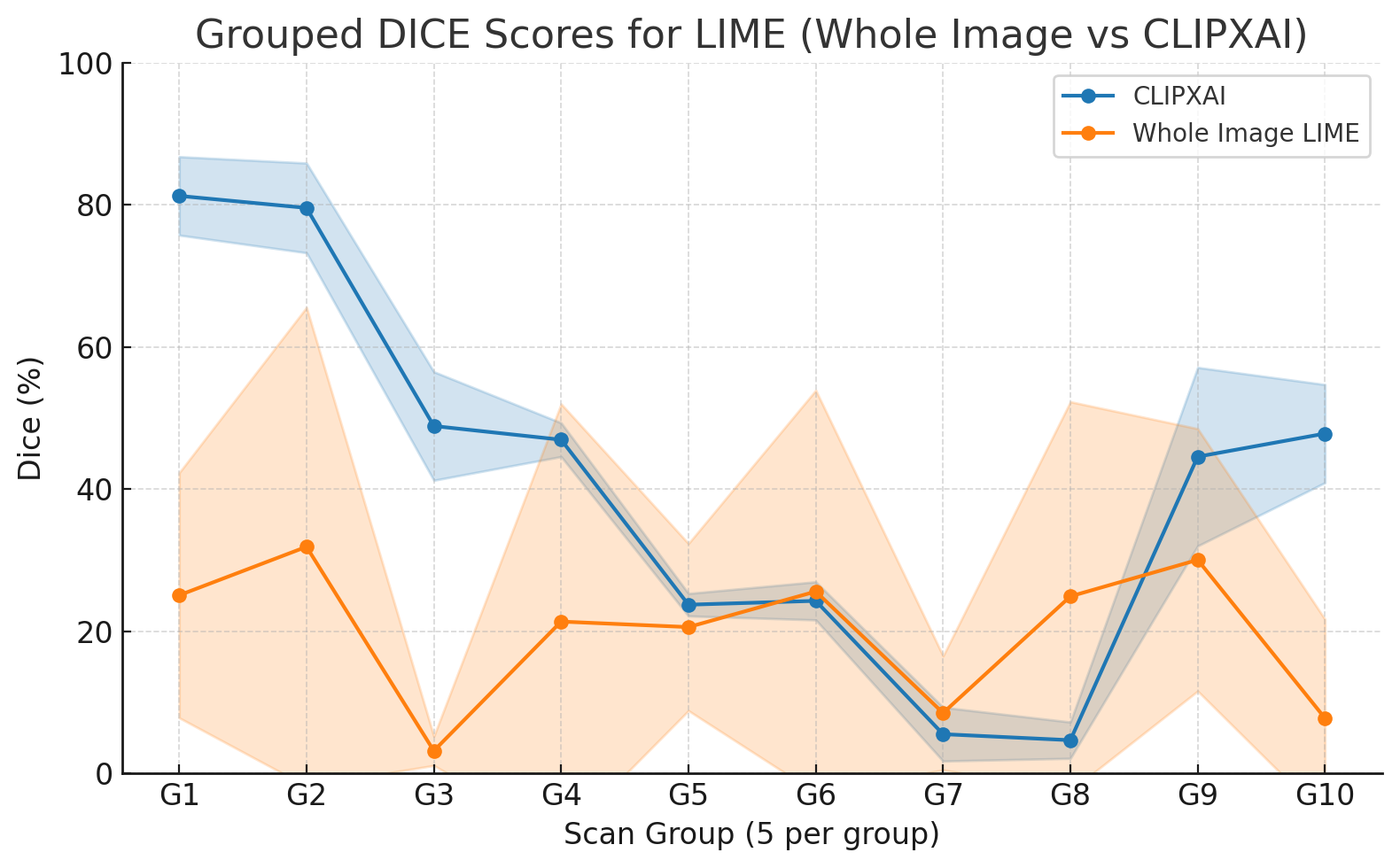}
    \vspace{0.5em}
    (a)
\end{minipage}
\hfill
\begin{minipage}{0.48\columnwidth}
    \centering
    \includegraphics[width=\linewidth]{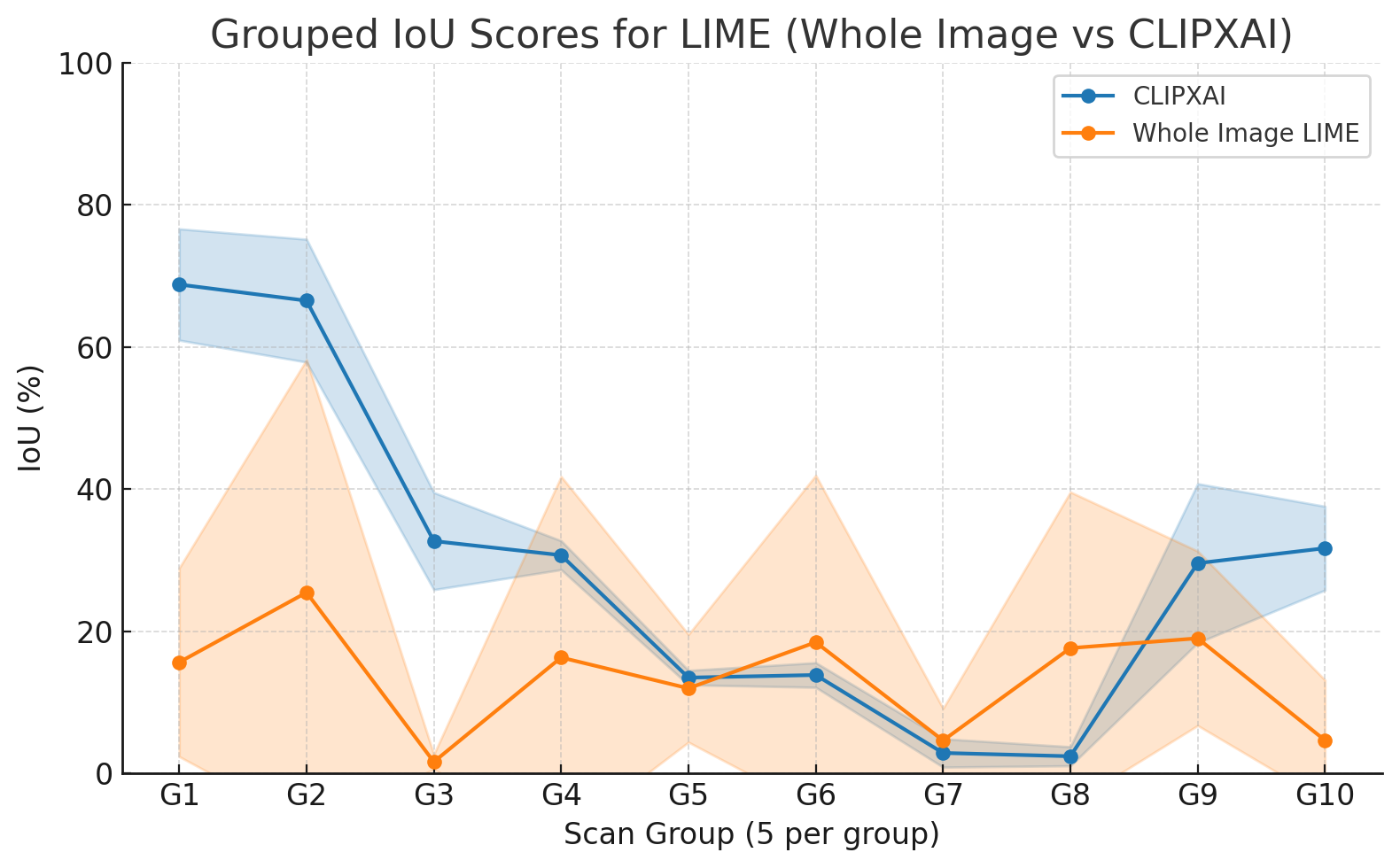}
    \vspace{0.5em}
    (b)
\end{minipage}

\caption{(a) Distribution of Dice values for Traditional LIME vs.\ XAI-CLIP. (b) Distribution of IoU values for Traditional LIME vs.\ XAI-CLIP.}
\label{fig:two_images_side_by_sideLIME}
\end{figure}

This is worth noting that the LIME-based analysis exhibited notable performance variability, with Dice and IoU values spanning from below 0.1 to as high as 0.893 and 0.802, respectively. In contrast, the traditional LIME baseline frequently failed to localize the target organs altogether, yielding dice scores close to zero in numerous test cases as shown in Fig. \ref{fig:two_images_side_by_sideLIME}, highlighting the limitations of unguided perturbation in complex anatomical settings.
\begin{table*}[t!]
\centering
\caption{Traditional vs.\ XAI-CLIP-guided explainability. $\uparrow$ indicates higher is better; $\downarrow$ indicates lower is better. Gray rows report the relative change of each XAI-CLIP variant compared to its traditional counterpart.}
\label{tab:clip_comparison}
\footnotesize
\resizebox{0.9\linewidth}{!}{%
\begin{tabular}{lccccc}
\hline
Method & Dice $\uparrow$ & IoU $\uparrow$ & GFLOPs $\downarrow$ & Time A $\downarrow$ & Time B $\downarrow$ \\
\hline
Traditional Occlusion & 0.6825 & 0.4956 & 168{,}780 & 401 & 308.5 \\
XAI-CLIP Occlusion    & 0.9871 & 0.9748 & 146{,}432 & 151 & 94.4 \\
\rowcolor[gray]{0.9}
$\Delta_{\text{XAI-CLIP}}$ & +44.6\% & +96.7\% & +13.2\% & +62.3\% & +69.4\% \\
\hline
Traditional LIME & $>0.1$ & $>0.1$ & 131{,}921 & 420 & 283.7 \\
XAI-CLIP LIME    & 0.412  & 0.297  & 94{,}431  & 311 & 223.7 \\
\rowcolor[gray]{0.9}
$\Delta_{\text{XAI-CLIP}}$ & --- & --- & +28.4\% & +26.0\% & +21.1\% \\
\hline
Traditional RISE & $>0.1$ & $>0.1$ & 181{,}010 & 1{,}040 & 533.3 \\
XAI-CLIP RISE    & $>0.1$ & $>0.1$ & 169{,}932 & 753 & 803 \\
\rowcolor[gray]{0.9}
$\Delta_{\text{XAI-CLIP}}$ & --- & --- & +6.1\% & +27.6\% & +33.6\% \\
\hline
\end{tabular}%
}
\end{table*}

\begin{figure}[t!]
\centering
\includegraphics[width=\columnwidth]{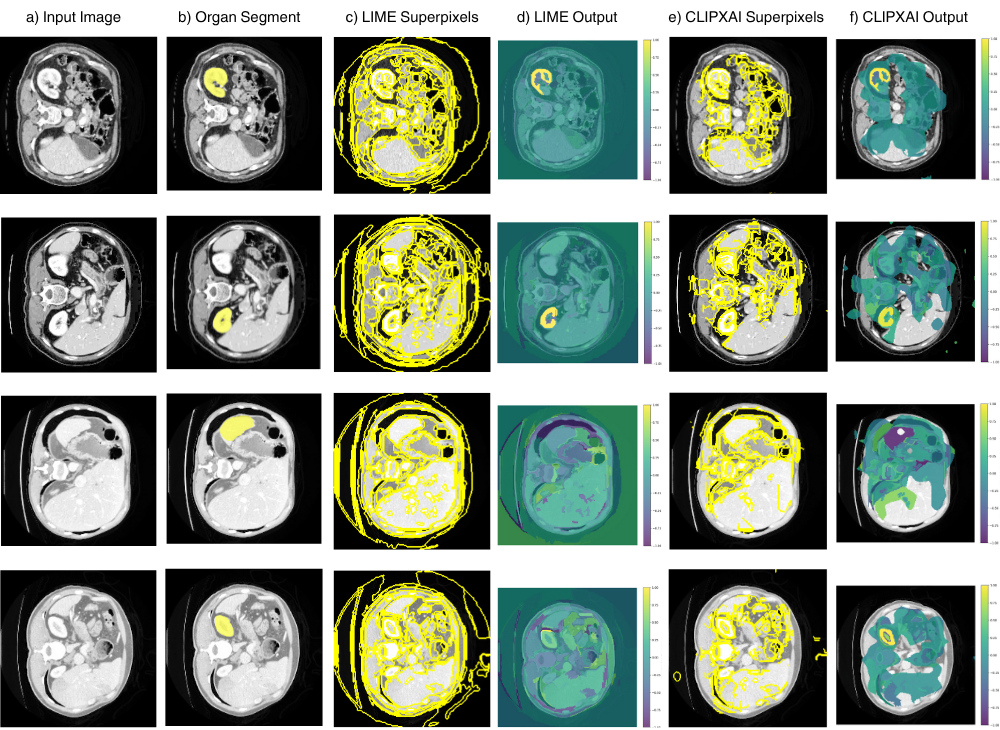}
\caption{LIME visualizations across different processing stages (a–f).}
\label{fig:lime_6columns}
\end{figure}

Notably, XAI-CLIP-LIME successfully identified anatomical structures even in scenarios where the baseline method completely failed. In instances where both methods achieved reasonable localization, XAI-CLIP-LIME maintained significant heatmap quality, as shown in Fig \ref{fig:lime_6columns}, while simultaneously reducing computational requirements by $28\%$ in GFLOPs and approximately $25\%$ in runtime. However, the consistently poor baseline performance (below $0.1$ in many cases) precluded meaningful percentage improvement calculations for this method.

\begin{figure}[t!]
\centering
\setlength{\tabcolsep}{0pt} % no horizontal gap
\renewcommand{\arraystretch}{0} % no vertical padding

\begin{tabular}{@{}c c c c@{}}
 % Reduced to 0pt

\includegraphics[width=\columnwidth]{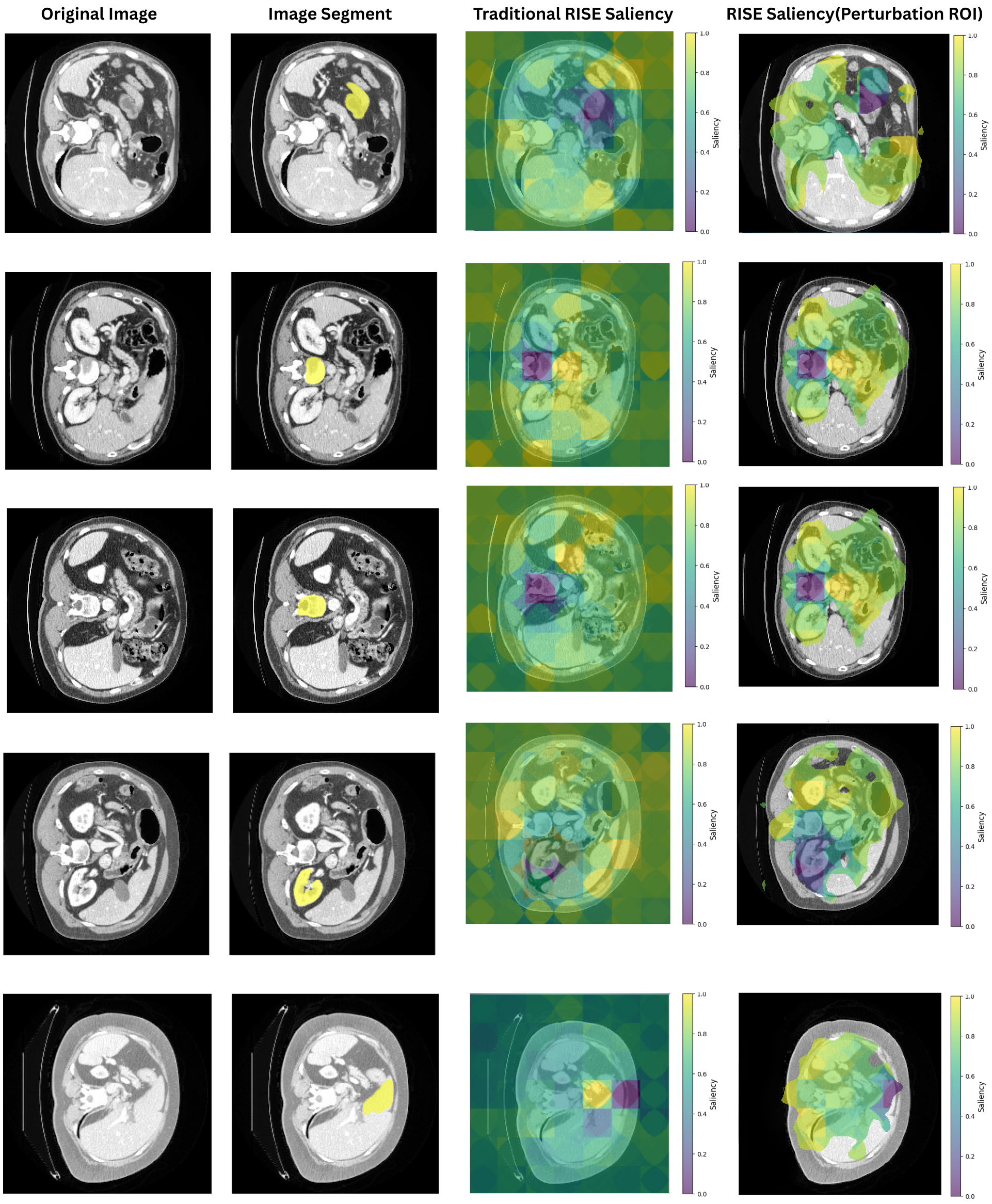}

\end{tabular}

\caption{RISE visualizations across different processing stages (a–d).}
\label{fig:rise_6columns}
\end{figure}

% RISE analysis revealed an unexpected yet consistent pattern in which the \emph{darkest} regions (lowest importance) coincided with the true organ locations, while importance gradually increased toward surrounding background tissue. Consequently, the saliency masks focused on neighbouring structures and background rather than the target organ, yielding Dice and IoU scores below $0.1$ for both baseline and XAI-CLIP variants due to minimal overlap with ground-truth boundaries. Traditional RISE occasionally produced random or image-wide highlights, whereas \mbox{XAI-CLIP-RISE} preserved the same counter-intuitive trend but in a more stable fashion and delivered a runtime reduction of $287$ s (a $28\%$ improvement).
Similarly, RISE-based analysis revealed a counterintuitive yet consistently observed behavior in which regions assigned the lowest importance, corresponding to the darkest areas in the saliency maps, aligned with the true organ locations, while importance progressively increased toward surrounding background tissue as demonstrated in Fig \ref{fig:rise_6columns}. As a result, the generated saliency masks emphasized neighboring structures and background regions rather than the target organs, leading to dice and IoU scores below 0.1 for both the baseline and XAI-CLIP variants due to minimal overlap with ground-truth boundaries. While traditional RISE occasionally produced unstable or image-wide attributions, XAI-CLIP-RISE preserved the same inverted importance trend in a more stable and structured manner, while also achieving a runtime reduction of 287 seconds, corresponding to a 28\% improvement.

In summary, we note that while the dice coefficient measures segmentation accuracy through spatial overlap with ground-truth masks, it is inherently sensitive to organ size, which accounts for the lower dice scores observed for smaller structures such as the spleen and pancreas. In contrast, AUC-ROC evaluates the model’s ability to assign higher confidence to true organ pixels relative to background without thresholding probabilistic outputs, making it particularly informative for assessing region-level confidence even when boundary delineation is imperfect. The computational efficiency gains achieved by the proposed framework arise directly from restricting perturbations to anatomically relevant regions, thereby avoiding unnecessary evaluations over background areas. Notably, the observed behavior in RISE-based explanations suggests that the segmentation model leverages contextual and boundary information surrounding organs more heavily than intra-organ texture cues; consequently, masking interior regions has limited impact on predictions, leading to low attributed importance for clinically relevant interiors despite their diagnostic significance.

\section{Conclusions}

% Explainable AI  remains a critical and indispensable component of medical image segmentation. In this work, we introduced XAI-CLIP, a novel framework that integrates CLIP-based vision language modeling, prompt-based learning, and segmentation models to enhance perturbation-based explainability. Our approach improves the interpretability and clarity of model explanations while substantially reducing the computational overhead typically associated with perturbation-based methods. Future work will focus on further optimizing the efficiency of XAI-CLIP and expanding its evaluation to real-world clinical datasets.
Explainable artificial intelligence remains a fundamental requirement for the safe and trustworthy deployment of medical image segmentation systems in clinical practice. In this work, we presented XAI-CLIP, a region-guided explainability framework that integrates vision-language representations, prompt-based learning, and segmentation models to enhance perturbation-based explanations. By leveraging semantically informed region localization, the proposed approach produces clearer, anatomically meaningful attribution maps while significantly reducing the computational cost associated with conventional perturbation strategies. Experimental results demonstrate that XAI-CLIP improves both interpretability and efficiency across multiple explainability methods and datasets. Future work will explore further optimization of the framework and its validation on large-scale, real-world clinical datasets to support robust and clinically deployable explainable medical imaging systems.

% \section*{Acknowledgment}

% The preferred spelling of the word ``acknowledgment'' in American English is 
% without an ``e'' after the ``g.'' Use the singular heading even if you have 
% many acknowledgments. Avoid expressions such as ``One of us (S.B.A.) would 
% like to thank $\ldots$ .'' Instead, write ``F. A. Author thanks $\ldots$ .'' In most 
% cases, sponsor and financial support acknowledgments are placed in the 
% unnumbered footnote on the first page, not here.

\section*{References}
\bibliographystyle{ieeetr}
\bibliography{refrences}

% \begin{IEEEbiographynophoto}{Second B. Author,} photograph and biography not available at the
% time of publication.
% \end{IEEEbiographynophoto}

% \begin{IEEEbiographynophoto}{Third C. Author Jr.} (Member, IEEE), photograph and biography not available at the
% time of publication.
% \end{IEEEbiographynophoto}

\end{document}